**Human vs Large Language Models: Judgmental Forecasting in an Era of Advanced AI**

Mahdi Abolghasemi

School of Mathematics and Physics

The University of Queensland

m.abolghasemi@uq.edu.au

Odkhishig Ganbold

Royal Melbourne Hospital Department of Medicine, Melbourne Medical School,

Faulty of Medicine, Dentistry and Health Sciences

The University of Melbourne

odkhishig.ganbold@unimelb.edu.au

Kristian Rotaru

Department of Accounting,  Monash Business School,  Monash University

BrainPark, Turner Institute for Brain and Mental Health, School of Psychological Sciences and Monash Biomedical Imaging Facility, Monash University

kristian.rotaru@monash.edu

**ABSTRACT**

This study investigates the forecasting accuracy of human experts versus Large Language Models (LLMs) in the retail sector, particularly during standard and promotional sales periods. Utilizing a controlled experimental setup with 123 human forecasters and five LLMs, including ChatGPT-4, ChatGPT3.5, Bard, Bing, and Llama2, we evaluated forecasting precision through Absolute Percentage Error. Our analysis centered on the effect of the following factors on forecasters' performance: the supporting statistical model (baseline and advanced), whether the product was on promotion, and the nature of external impact. The findings indicate that LLMs do not consistently outperform humans in forecasting accuracy and that advanced statistical forecasting models do not uniformly enhance the performance of either human forecasters or LLMs. Both human and LLM forecasters exhibited increased forecasting errors, particularly during promotional periods. Our findings call for careful consideration when integrating LLMs into practical forecasting processes.

## 1. Introduction

Retail supply chain optimization and inventory management critically depend on the precision of demand forecasting. Effective forecasting integrates two principal components: quantitative models and expert human judgment (Brau et al., 2023, Perera et al., 2019). Quantitative forecasting has achieved a level of sophistication and is systematically implemented in various forecasting solutions; however, the nuance of human judgment remains less quantifiable and is not as thoroughly researched (Write et al., 1996; Zellner et al., 2021; Sroginis et al., 2023). Human expertise plays a pivotal role at several points in the forecasting continuum – ranging from data selection to model optimization, and most notably, in the application of discretionary adjustments to model outputs (Arvan et al., 2019; Lawrence et al., 2006; Perera et al., 2019). Such expert-informed adjustments are suggested to improve forecast accuracy by integrating qualitative market insights with quantitative predictions – insights often elusive to purely algorithmic approaches (Fildes et al., 2009; Lawrence et al., 2006). However, there have been limited efforts to systematically explore in controlled experimental settings the mechanisms and reasoning behind experts' modifications of forecasts, and specifically how the integration of advanced forecasting methods and access to qualitative information about past and upcoming promotional periods may influence the precision of expert-informed adjustments (Fildes et al., 2019; Sroginis et al., 2023).

The integration of generative AI and Large Language Models (LLMs) into forecasting represents a notable development, directly following the above discussion on the role of human expertise in enhancing statistical forecasts. LLMs are complex computational models trained on large-scale datasets, capable of generating remarkably human-like text (Brown et al., 2020; Chang et al., 2023). As LLMs like GPT can analyze vast amounts of qualitative information well beyond the capabilities of human forecasters, integrating LLMs into forecasting processes will likely enhance performance in the near future (Kirshner, 2024). Their ability to process and analyze large volumes of data and provide, contextually relevant information has sparked interest in their potential to enhance judgmental forecasting (Makridakis et al., 2023; Schoenegger et al., 2024a). Such AI models are equipped to absorb and recall information, propose actionable insights, and engage in reasoning processes that may rival human cognitive abilities (Yu et al., 2023; Xu et al., 2022). Despite their potential, the performance of LLMs in judgmental forecasting, particularly during promotional periods characterized by complex endogenous and exogenous variables, is not well established (Makridakis et al., 2023). Consequently, the comparison of forecasting performance between LLMs and human forecasters emerges as an important research question (Schoenegger et al., 2024a,b; Halawi et al., 2024; Kirshner, 2024).

Recent studies in the field of forecasting indicate that LLMs could provide valuable predictive insights for economic and market forecasting (Yu et al., 2023; Zhao et al., 2024). These models have shown a level of understanding and capability that is noteworthy in these complex domains. A study by Li et al. (2023) specifically explores the potential of ChatGPT as a financial advisor, examining its ability to forecast listed firm performance. Their findings suggest that ChatGPT can correct the optimistic biases of human analysts, demonstrating the potential of LLMs in offering more balanced and accurate forecasts in financial contexts. This aligns with the broader trend of employing advanced analytical methods with LLMs, potentially enhancing the forecasting process by leveraging their extensive knowledge base and inferential reasoning. However, recent empirical evidence also highlights certain limitations of LLMs compared to human forecasters, particularly in tasks that require probabilistic forecasting of future events. Notably, Schoenegger and Park (2023) found that in a real-world forecasting tournament, GPT-4's probabilistic forecasts were significantly less accurate than those made by human participants. This underperformance of LLMs in scenarios requiring advanced decision-making and adaptive reasoning suggests that while LLMs like ChatGPT can offer advantages in specific areas, such as mitigating human biases in financial forecasting, their capabilities in more dynamic and uncertain forecasting scenarios remain limited.

This study aims to methodically assess the inherent capabilities of LLMs in judgmental forecasting and to compare their performance with that of human experts. In controlled experiments, both LLMs and human forecasters operate without any prior training or feedback on forecast accuracy, following the methodology used in studies such as Schoenegger et al. (2024b). Consequently, our research directly evaluates the default capabilities of LLMs, providing an unadulterated assessment of their judgment and decision-making processes under experimental conditions. We explore the performance of LLMs across various scenarios, including their use alongside both simple baseline and advanced statistical models, their effectiveness during sales promotions, and their responses to external variables that add complexity to forecasting tasks.

By calculating and comparing the Absolute Percentage Error (APE) of forecasts produced by human experts and a selection of LLMs, including the advanced ChatGPT4, we aim to contribute new insights into the efficacy of AI-driven models in enhancing forecasting accuracy. This method aligns with the approach advocated by Sroginis, Fildes, and Kourentzes (2023), who support the use of controlled, laboratory-based experiments as a means to obtain an accurate assessment of judgmental adjustments in the forecasting process. Indeed, laboratory experiments have been instrumental in demonstrating the importance of contextual understanding and product-specific knowledge, which extend the accuracy of forecasts beyond the capabilities of statistical models alone (Sroginis, Fildes, and Kourentzes, 2023, Arvan et al., 2019; Webby et al., 2005). By doing so, we aim to provide a balanced view of the forecasting capabilities across both human and machine-based forecasters, offering a better understanding of the strengths and limitations inherent to each approach considering the current level of development of LLMs.

The findings of our study present a comparison between human forecasters and LLMs in terms of their forecasting accuracy measured by APE. Analysis indicates that LLMs do not uniformly exceed the predictive accuracy of human forecasters. Specifically, the most recent models such as ChatGPT4 and Bing have shown a level of performance comparable to that of humans, whereas earlier models including ChatGPT3.5, Bard, and Llama2, register higher APE. This suggests a variability in the ability of different LLMs to process large datasets and produce accurate forecasts, possibly due to unique model limitations or the need for more sophisticated training to cope with diverse sales environments. A closer examination of the data indicates varying degrees of response to sales data, forecast type, promotional periods, and external factors, by both human forecasters and LLMs.

This research enriches the forecasting literature by providing an analysis that compares the performance of human forecasters with that of LLMs across diverse conditions, emphasizing the sophistication of forecasting models, the sales context, and external influences as critical factors in performance assessment. The study examines the capacity of LLMs to interpret information amidst promotions and external variables, offering a comprehensive evaluation of these models' roles in complex forecasting situations. The outcomes of this research have substantial implications for the enhancement of demand planning and the design of advanced forecasting support systems that more seamlessly incorporate human expertise. Furthermore, our work opens new avenues for research into the integration of LLMs with existing forecasting practices and the refinement of collaborative human-AI decision support frameworks.

The remainder of this paper is organized as follows: Section 2 reviews relevant literature and formulates the hypotheses; Section 3 outlines the methodology and describes the participants; Section 4 presents the findings; Section 5 discusses these results and concludes with a reflection on the study's limitations and directions for future research.

**2. Background and hypothesis development**

*2.1. Background and literature review*

In sales forecasting, the integration of statistical models, advanced machine learning algorithms, and human judgment has become increasingly sophisticated. The evolution of forecasting models from simple baseline to advanced integrated systems reflects this complexity. Baseline models, effective at capturing fundamental trends, often struggle in volatile scenarios like promotions (De Baets & Harvey, 2018). Advanced models, incorporating techniques such as dynamic linear regression, provide a robust solution by integrating complex factors like promotional impacts and pricing strategies (Ma, Fildes, & Hung, 2016

Kourentzes & Petropoulos, 2016; Abolghasemi et al., 2020b). This progression addresses the cognitive challenges forecasters face and highlights the need for balancing statistical rigor with human judgment (Fildes & Goodwin, 2007; Fildes et al., 2009).

Human judgment, essential in adjusting statistical forecasts, particularly in promotional contexts, introduces complexities and biases impacting forecast quality (Goodwin & Wright, 2010; Sroginis et al., 2023). In promotional scenarios, the complexity and unpredictability inherent to consumer behavior necessitate a judicious combination of model-based forecasts and expert intuition (Fildes & Hastings, 1994; Sroginis et al., 2023). The significance of human judgment in forecasting, particularly in scenarios influenced by promotions, represents a crucial field of inquiry in both academic research and practical application. The effectiveness of human intervention in these settings exhibits considerable variability, as evidenced by studies such as Kourentzes and Fildes (2023), Trapero et al. (2013), Fildes and Goodwin 2007).

The cognitive load associated with complex forecasts can result in suboptimal decision-making, highlighting the need for more intuitive and supportive forecasting systems (Lim & O'Connor, 1995; Lawrence et al., 2006). The literature shows that cognitive biases and heuristics significantly influence forecasters' judgments, often leading to systematic errors (Fildes, Ma, & Kolassa, 2022; Harvey, 1995; Bolger & Harvey, 1993; Goodwin & Wright, 2010). Thus, the trajectory of current research is steering towards a more integrated approach, where the strengths of human cognition are leveraged to complement and refine the outputs generated by advanced forecasting models (Sroginis et al., 2023; Kouentzes and Fildes, 2023). However, further exploration into the balance between human judgment and statistical models is necessitated to better understand the value added by human intervention.

The evolution from purely statistical to judgmental forecasting marks a shift towards integrating historical data with the context of current scenarios, aligning with the core objective of forecasting support systems (FSS). These systems aim to harmonize algorithm precision with human intuition to enhance overall forecast accuracy (Fildes & Goodwin, 2007; Zhang et al., 1998; Green & Armstrong, 2015). However, there remains a gap in understanding the rationale behind judgmental adjustments, particularly the impact of special events, biases, and human factors on judgmental forecasting (Bolger & Önkal-Atay, 2004; Fildes & Goodwin, 2007; Lawrence et al., 2006; Sroginis et al, 2023). This gap calls for a deeper investigation into the cognitive processes involved in judgmental forecasting. Fildes et al. (2009) have emphasized the complexity of these processes, revealing how biases and heuristics can influence forecasters' judgments. Bolger and Harvey (1993) explored how cognitive biases can lead to systematic errors, highlighting the need for systems that support and improve human judgment. Harvey (1995) emphasized the importance of understanding how individuals process and interpret information in forecasting. Goodwin and Wright (2010) showed how cognitive biases can influence judgmental forecasting, especially in complex or ambiguous situations. In essence, the progression towards integrating statistical and judgmental forecasting demands advanced analytical tools and a comprehensive understanding of the human element in forecasting.

The interaction between human cognition and statistical data needs further exploration to enhance forecast efficacy and accuracy, particularly in complex and uncertain environments. Our investigation explores both simple baseline forecasts and advanced promotional forecasts. Simple baseline forecasts effectively capture underlying time series trends and seasonality, yet frequently fail to account for volatile elements like promotions (De Baets & Harvey, 2018). In contrast, advanced baseline forecasts utilize dynamic linear regression models. These models are designed to address the need for a more robust forecasting tool that can assimilate the effects of promotional impacts and integrate key variables such as pricing strategies (Kourentzes & Petropoulos, 2016; Abolghasemi et al., 2020b). The shift towards advanced forecasting models is a response to forecasters' limited cognitive resources. These models aim to reduce cognitive load and provide a more strategic overview of data for effective decision-making (Bolger & Harvey, 1993; Fildes et al., 2009; Fildes, 1991). The role of advanced models in mitigating the biases inherent in human judgment demonstrates their potential in enhancing the accuracy and reliability of forecasts. The clarity provided by these models can be crucial in avoiding the pitfalls of heuristic-driven decisions (Harvey and Bolger, 1996), which are more likely under conditions of excessive cognitive load. Thus, our exploration into the dichotomy of simple baseline versus advanced promotional forecasting models extends beyond a mere a comparative performance assessment; it is also an investigation into their potential to enhance human analytical capabilities and judgment in forecasting scenarios characterized by complexities, such as promotional periods.

The advent of LLMs like ChatGPT adds a new dimension to this research gap. Trained on extensive datasets and textual information, these models generate detailed, contextually relevant information, with potential applications in economic and market forecasting still under investigation (Schoenegger & Park, 2023; Makridakis et al., 2023). Their performance in complex scenarios, especially in promotional periods, is an emerging area of research. It is thus important to understand whether LLMs, as a new class of models, change dramatically our knowledge of the performance of algorithmic models in judgmental forecasting or, put differently, whether LLMs outperform human forecaster, considering a variety of critical factors that have been considered in judgmental forecasting literature. Below we develop our theoretical predictions to contribute further to this investigation.

*2.2. Hypothesis development*

The study's hypotheses are directed towards evaluating the performance of LLMs relative to human judgment under various conditions, including simple baseline vs advanced forecasts, promotional impacts, and external impact variables. This approach aligns with research emphasizing the complex dynamics between algorithmic models and human cognitive processes (Lawrence et al., 2006; Goodwin & Wright, 2010). We also propose hypotheses concerning the forecasting accuracy of humans and five distinct LLMs (assessing each forecaster individually) under these diverse circumstances. To address the potential confounding effects of actual sales figures on our forecast accuracy metric (i.e., absolute percentage error), we have incorporated actual sales as a control variable in our analysis. This adjustment is crucial for mitigating the potential confounding effects of fluctuations in sales figures, which can skew the performance metrics of forecasters. By controlling for actual sales, we aim to isolate the true impact of the variables under investigation, providing a more precise estimation of how various conditions directly affect forecasting accuracy. This ensures that our results more accurately reflect the specific effects of the variables of interest, rather than being influenced by the actual sales volatility.

*2.2.1 Comparative Accuracy of LLMs and Human Forecasters (H1)*

The adoption of LLMs in time series forecasting prompts a critical evaluation of their performance compared to human forecasters. Human forecasters integrate a broad spectrum of experiential and contextual knowledge into their predictions and often employ heuristics to simplify complex tasks. However, their judgments are vulnerable to several well-documented cognitive biases such as overconfidence, anchoring, and confirmation biases, which impair the accuracy of their forecasts (Kahneman & Tversky, 1974; Chapman & Johnson, 1999; Lawrence et al., 2006). LLMs, on the other hand, leverage vast amounts of data and sophisticated pattern recognition algorithms to generate forecasts, which may enhance consistency and reduce susceptibility to these biases. Nevertheless, LLMs can also manifest new biases such as hallucinations, which adversely impact their performance (Silver et al., 2016; Brown et al., 2020).

In the domain of sales forecasting, achieving accuracy is crucial, and the ongoing debate pits human expertise against algorithmic precision. Human forecasters bring important insights into market dynamics and can swiftly adapt to new information, but their effectiveness is often hampered by psychological biases and a limited capacity to process large data volumes (Lawrence et al., 2006). In contrast, LLMs, unencumbered by these psychological limitations and adept at processing extensive datasets, present a promising alternative. With their ability to discern trends and patterns from comprehensive data (Silver et al., 2016), LLMs are hypothesized to yield more accurate judgmental forecasts than human forecasters, especially in complex forecasting scenarios such as promotional sales. This hypothesis anticipates a discernible difference in forecasting performance, favoring LLMs over humans, in terms of accuracy.

***Hypothesis 1 (H1):*** *LLMs will surpass human forecasters in forecasting accuracy, when adjusted for comparable levels of actual sales.*

2.2.2 Enhanced Forecasting Performance with Advanced Forecast Models (H2)

In exploring the impact of advanced promotional models on forecasting accuracy (e.g. Sroginis et al., 2023), it is important to consider how additional information in the form of expert predictions or algorithmic projections may influence the decision-making processes of both human forecasters and LLMs. Research in judgmental forecasting suggests that humans often benefit from having access to advanced forecast information, which can serve as an anchor and potentially improve accuracy by reducing the range of possible outcomes considered (Lawrence et al., 2006; Goodwin & Fileds, 2022). However, humans may also over-rely on this information and underweight their own judgment (Meub & Proeger, 2016).

For LLMs, advanced forecasting models can provide accurate data to be integrated into the LLMs' reasoning process, potentially enriching the model's inputs and leading to better performance, especially when these advanced forecasts encapsulate complex patterns that the simple baseline models alone might not detect. Nevertheless, the effectiveness of this integration depends on the quality of the forecasts and the ability of the LLMs to appropriately weigh this information against the data it has already processed.

As discussed in our literature review and in the lead to Hypothesis 1, the provision of advanced forecasts is likely to influence the accuracy of subsequent predictions made by both human forecasters and LLMs. For human forecasters, previous literature has highlighted how the use of advanced forecasts can serve as a heuristic, aiding in the process of making complex judgments and potentially leading to improved accuracy, though it also carries the risk of over-reliance and the suppression of independent judgment (Goodwin, 2005; Lawrence et al., 2006). On the other hand, for LLMs, which are designed to synthesize large volumes of data and identify patterns, advanced forecasts could provide an additional layer of information, enhancing the model's predictive capabilities (Makridakis et al., 2023). The advanced forecasts, if accurately capturing future trends, could therefore improve the LLM's predictive performance by offering a form of 'expert input' that the model can incorporate into its algorithms effectively. This integration is not influenced by the psychological biases that affect human forecasters, potentially leading to a more optimal use of the provided advanced forecasts.

Given the above considerations, we hypothesize:

***Hypothesis 2 (H2):*** *Both human forecasters and LLMs will exhibit improved forecasting performance when provided with advanced forecasts, due to the additional information serving as a data-rich input that can aid in refining predictions (H2a). However, the degree of improvement will differ between the two* *groups, with LLMs expected to utilize advanced forecasts more effectively, as they can systematically integrate and evaluate such inputs without the cognitive biases that human forecasters are subject to (H2b).*

*2.2.3 Variability in Forecast Accuracy During Promotional and Non-Promotional Periods (H3)*

Building upon the challenges posed by promotional periods in sales forecasting, our theoretical prediction hinges on the systematic data processing capabilities of LLMs. LLMs' proficiency in forecasting during promotional periods, however, is predicated on their exposure to a diverse set of historical promotional data. Such data enable these models to adjust to the volatility introduced by promotions more consistently than humans, who might underreact or overreact to the influence of promotions on consumer behavior. While LLMs are not explicitly trained on promotional period data in this study, their inherent capabilities may already include the ability to anticipate the impact of promotional activities on sales. This potential understanding could result in more consistent forecasting accuracy across promotional and non-promotional periods compared to human forecasters, as we test our predictions.

Thus, Hypothesis 3 is formulated with the expectation that LLMs, given their algorithmic nature and assuming comprehensive training on promotional data, will demonstrate a less pronounced difference in forecasting accuracy between promotional and non-promotional periods compared to human forecasters, whose judgment may be clouded by inherent cognitive biases:

**Hypothesis 3 (H3):** *The accuracy of both human forecasters and LLMs will differ during promotional versus non-promotional periods (H3a), with LLMs expected to demonstrate a less pronounced difference in accuracy between these periods compared to humans (H3b).*

*2.2.4 Impact of External Conditions on Forecasting Disparities (H4)*

Forecasting in the face of external impacts is a complex endeavor, requiring the consideration of diverse, often non-linear influences on sales data. Understanding the influence of external conditions on forecasting accuracy necessitates distinguishing between negative and positive impacts. In our study, external impact is presented as a compound effect of competitors' activity such as advertisement, indicating whether such activity has impacted or will impact product sales positively or negatively. These conditions introduce volatility into sales data, complicating the forecasting process (Abolghasemi et al., 2020a).

Human forecasters, when integrating and interpreting external impact information – often qualitative in nature – can exhibit significant biases influenced by the emotional valence of the information, whether negative or positive (Sroginis et al., 2023). During economic downturns or natural disasters, for instance, forecasters might disproportionally emphasize such negative information, resulting in overly conservative estimates due to a well-documented negativity bias (Lewandowsky et al., 2012; Ito et al., 1998).

Conversely, during periods of positive economic outlooks or favorable holiday seasons, forecasters may demonstrate undue optimism in their predictions (Eroglu & Croxton, 2010; Goodwin & Fildes, 2022; Önkal et al., 2023). Negative information often influences evaluations more strongly than comparably significant positive information and is a recognized phenomenon in expert forecasts (Chang & Hao, 2022). Such asymmetry in human judgment, often driven by affect and representativeness heuristics, leads to greater variability in forecasting accuracy, particularly during periods of negative versus positive external impact.

In contrast, LLMs, devoid of emotional responses, analyze data patterns algorithmically. Provided their default training includes a range of both negative and positive external impacts, LLMs can maintain a consistent forecasting approach. They adjust to different conditions based on learned data patterns rather than emotional responses, a capability rooted in their algorithmic nature (Chang et al., 2023). Therefore, while both human forecasters and LLMs are likely to experience variances in forecasting accuracy due to external impacts, the extent of these variances may differ. Humans are expected to show higher sensitivity to the valence of external information due to emotional and cognitive biases. LLMs, unaffected by such biases, are anticipated to exhibit a smaller disparity in APE across varying external conditions.

***Hypothesis 4 (H4):*** *The accuracy of both human forecasters and LLMs will differ under negative versus positive external impact conditions (H4a). Human forecasters will show a greater difference in accuracy under negative versus positive external impact conditions (i.e., lower accuracy) than LLMs (H4b).*

*2.2.4 Variability in Forecast Accuracy under Complex Forecasting Scenarios (H5)*

The dynamic environment of sales forecasting, where numerous factors intersect, may impose challenges for human forecasters to maintain high levels of accuracy due to the cognitive strain imposed by the need to balance and interpret multiple, and sometimes conflicting, pieces of information. The propensity for cognitive biases and the potential for information overload may lead to increased variability in accuracy among human forecasters when considering all factors simultaneously.

When considering a comprehensive forecasting model that accounts for all previously considered factors – such as the presence of advanced forecasts, promotional versus non-promotional periods, and negative versus positive external impacts (forming therefore six treatment conditions, see Table 1) – the interaction between these variables becomes crucial in determining forecasting accuracy. This multifaceted approach reflects the real-world complexity of sales forecasting, where multiple factors often simultaneously influence outcomes.

**Table 1.** Treatment conditions tested in this study.

| Treatment condition | Promotion | Statistical Model | External Impact |
|---|---|---|---|
| 1 | Yes | Baseline | Positive |
| 2 | Yes | Advanced (promotional) | Positive |
| 3 | Yes | Baseline | Negative |
| 4 | Yes | Advanced (promotional) | Negative |
| 5 | No | Baseline | Positive |
| 6 | No | Baseline | Negative |

LLMs are designed to handle large volumes of complex data, potentially allowing them to better navigate the interactions between multiple forecasting factors without succumbing to the biases that affect human judgment (Chang et al., 2023). Their algorithmic efficiency and immunity to psychological biases allow them to systematically analyze complex interactions between factors. Therefore, when all factors from the forecasting model are taken into account, LLMs are hypothesized to achieve a more consistent and accurate performance compared to human forecasters, as indicated by a generally lower forecasting accuracy that does not fluctuate as widely with the change in conditions.

***Hypothesis 5:*** *The accuracy of human forecasters will vary across treatment conditions that incorporate advanced forecasts, promotional conditions, and external impacts (H5a). When considering a model that incorporates the aforementioned treatment conditions, LLMs will demonstrate more accurate forecasting across these conditions, as indicated by a consistently lower accuracy, compared to humans (H5b).*

**3. Method**

*3.1 Participants*

The experimental study has been conducted in August 2023 with one hundred twenty-three human forecasters and five LLMs. Human forecasters were business school graduate students (65.9% female, 34.1% male) who voluntarily participated in this study conducted in a behavioral laboratory at one of Australia's large universities. The average age of human forecasters was 24.21 (SD = 2.51), ranging from 22 to 34 years, and they had on average 1.46 years (SD = 1.60) of business experience. All human forecasters had practical skills in business forecasting since they all completed a unit on time series forecasting where they had to demonstrate practical skills in applying a variety of timeseries forecasting models for stationary and non-stationary data (including the data with additive or multiplicative seasonal effects).

In addition, we used five popular and powerful LLMs including ChatGPT3.5, ChatGPT4 developed by OpenAI, Llama2 developed by Meta, Bing developed by Microsoft, and Bard developed by Google. These models are the most advanced natural language processing models. They benefit from at least millions of parameters, have been trained on a vast amount of data, and fine-tuned to be able to reason, answer questions, and generate text like a human.

*3.2 Design*

The study's design was a 2x2+2 mixed design, involving three independent variables (IVs). The first IV, 'statistical forecast', had two conditions: a baseline and an advanced model. The second IV, 'external impact', differentiated between positive and negative effects. The third IV, 'promotion', was divided into active and no promotion conditions. Notably, the advanced statistical forecast, which accounted for promotional effects, was exclusively applied when promotions were active. Consequently, this design yielded six experimental conditions (as per Table 1):

1. Promotion active, positive external impact, baseline statistical forecast.
2. Promotion active, positive external impact, advanced statistical forecast.
3. Promotion active, negative external impact, baseline statistical forecast.
4. Promotion active, negative external impact, advanced statistical forecast.
5. No promotion, positive external impact, baseline statistical forecast.
6. No promotion, negative external impact, baseline statistical forecast.

Each participant evaluated the same collection of twenty-four time series. The accuracy of the participants' forecasts was gauged using the Absolute Percentage Error (APE) as the dependent variable (DV). APE quantifies forecasting precision by comparing the forecasted values against actual sales, presented as a percentage. The lower the APE value, the more accurate the forecast. This measure is critical for assessing the effectiveness of judgmental forecasting against statistical models under varying market conditions and promotional strategies.

Participants are awarded based on their performance (forecasting accuracy) between 5 to 20 AUD. The forecasting accuracy is calculated based on the APE as follows:

$$APE = \frac{abs(f_t - x_t)}{x_t} \quad (1)$$

where $f_t$ is the forecast at time $t$, and $x_t$ is the actual sales at time $t$.

*3.3 Materials and Apparatus*

The experiment was designed with *R shiny app* and hosted on *RStudio*[1]. Participants were given 24 different sales series representing the actual sales of different products in an Australian company. The company is a food manufacturing company that manufactures health and breakfast products and distribute their products through two main retailers in Australia. Each series has 24 observations with different scales of sales, promotion status, and sales uplifts due to promotion.

Figure 1 shows a screenshot of the experiment graphical interface. It represents the dashboard of the forecasting system that participants used to conduct their forecasting decisions.

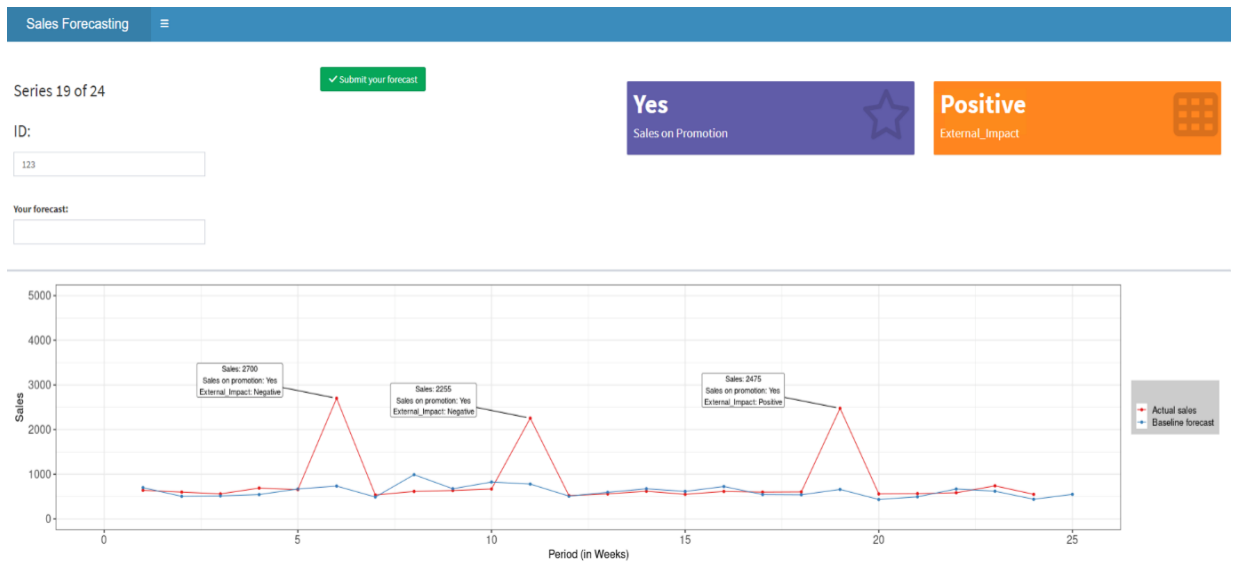

**Figure 1**: Interface of the experiment dashboard

"Promotion status" shown in the purple box at the top right of the dashboard is the main variable that causes sales uplifts. It is a binary variable that takes either "Yes" or "No" representing promotion presence and absence, respectively. The orange box represents the "External Impact" variable, a binary indicator summarizing the effects of competitors' activities on sales. It takes either "Positive" or "Negative" values to signify an increasing or decreasing impact on sales volume, respectively. This impact may vary between promotional and non-promotional periods and influences demand by marginally increasing or decreasing sales volumes up to 10%. This approach was influenced by a case study that examined how a company accounts for promotional impacts on sales while also considering competitor activities. The company's forecasts leverage continuous market intelligence, enabling specialists to adjust their predictions in response to comprehensive insights about competitors' policies, prices, and promotions. The influence of the "External Impact" variable on sales is dual: it is perceived negatively if competitors promote similar products when the company's products are not on major sale, and positively if the company's products are on major sale while competitors' products are not. While these impacts are specific to individual products, they are generalized by the company's experts based on overall market share and conditions. Although this model is derived from our case study, variations in other contexts might present worthwhile avenues for further research.

---

[1] The task can be accessed here: https://mahdiforecasting.shinyapps.io/exp1/

Besides this information, participants were given a "statistical forecast" depicted in "blue" color dotted line on the main dashboard. The statistical forecast may be of two types: "simple baseline statistical forecast" (aka 'baseline forecast') or "advanced statistical forecast" (aka 'advanced forecast'). The "simple baseline forecast" is a forecast generated by the forecast support system of the company, which uses "Holt-Winter exponential smoothing" model to forecast baseline sales in the absence of promotion and does not consider "promotion" and 'external impact'. Whereas the advanced forecast takes into account the impact of promotion but not 'external impact' and uses a "dynamic linear regression" model to forecast sales.

The given information includes two types of dotted lines, i.e. blue representing historical forecasts and red - historical sales for the last 24 periods of time series, depicted on the dashboard. We provided the same graph and dashboard for all participants to avoid any interfering impact of graph types (Ramires & Harvey, 2023). Other relevant information provided to the participants included: promotion status and external impact in Period 25, and system statistical forecast (Simple baseline or advanced baseline statistical forecast). We assume that all of the internal and external variables that contribute to sales are summarized in the given information.

*3.4 Procedures*

Participants (human forecasters and LLMs) were asked to assume the role of the sales manager in a hypothetical company. In this role, they were asked to forecast sales for Period 25 based on the given information. Each participant evaluated the same collection of twenty-four separate time series. These series represented genuine sales data for various health and breakfast products from an Australian food manufacturer, encompassing fluctuations in sales volume, promotional activities, and sales boosts from promotions. Period 25 is either promotion period or non-promotion period. For simplicity, participants were not asked to forecast any immediate post promotion periods.

The ability of LLMs to generate accurate responses heavily depends on the given prompts. Prompts are texts that human can give to LLMs to provide them with the context and ask them for responses. To ensure fairness in our experiment, we provided the LLMs with the same information that was given to the human participants. This was done during the same week in August 2023 when the experiment with the human forecasters was conducted. In our prompt engineering, the data provided effectively characterized the task for the LLMs. This information comprises the contents of handouts, time series observations in tabular format, and additional details such as promotion status, statistical forecasts, and external impact information, all of which were provided to the human participants. The precise prompt provided to the LLMs is detailed in Appendix B. The same prompt was used for all LLMs considered in this study. Detailed descriptive statistics for the experimental data are provided in Appendix C.

4**. Results**

We use both mixed-effects models and regression models (specifically for testing the performance of each LLM separately) to test our hypotheses. Adopting the approach of Fildes et al. (2019), our statistical analysis utilized linear mixed-effect models with restricted maximum likelihood, favored for handling correlated repeated-measures data and covariates that vary over time, due to their flexibility, resilience to missing data, and ability to model realistic variance and correlation patterns (Gueorguieva & Krystal, 2004; Cnaan, Laird, & Slasor, 1997).

The general equation for our mixed-effects models is:

$$Y_{ij}=\beta_0+\beta_1 X_{ij}+\beta_2 Z_{ij}+u_j+\epsilon_{ij}$$

Where:

- $Y_{ij}$ represents the dependent variable (APE) for the $i$-th observation within the j-th forecaster.
- $\beta_0$ is the intercept.
- $X_{ij}$ is the covariate for actual sales, reflecting its impact on APE.
- $\beta_1$ is the coefficient for actual sales.
- $Z_{ij}$ encompasses the categorical fixed effects from various experimental conditions, including forecaster type (humans and LLMs), forecast model (baseline or advanced), promotional status (promotion or non-promotion), and external impact (positive or negative), along with their interactions.
- $\beta_2$ are the coefficients for the fixed effects.
- $u_j$ is the random effect for each forecaster, capturing unexplained variability across forecasters.
- $\epsilon_{ij}$ is the residual error for each observation.

In our analysis, "actual sales" is included as a covariate to mitigate the influence of sales volume variations on APE. Experimental conditions, such as promotion status (yes promotion, no promotion), forecast type (baseline forecast, advanced forecast), external impact (positive external impact, negative external impact), and treatment interventions (interaction of aforementioned conditions), are integrated as fixed effects. This structure helps clarify how different forecasting contexts affect forecast accuracy. By treating each forecaster as a random effect, our model adapts to individual differences among forecasters, thus enhancing the analysis depth. This approach not only captures the variability inherent in real-world forecasting but also boosts the generalizability and reliability of our findings.

When testing hypotheses for each forecaster type, where each condition is represented by a single sample observation (i.e., each LLM forecaster), we employ regression models because the data structure does not support the testing for random effects, which are a requisite for the application of mixed-effect models. We discuss the main results in the following sections. More comprehensive results on H2a-H5a are detailed in Appendix J. Additionally, descriptive statistics of forecast accuracy for both human and LLM forecasters are provided in Appendices D and E.

*4.1 Comparative Accuracy of LLMs and Human Forecasters (H1)*

Our Hypothesis 1 posited that LLMs would surpass human forecasters in terms of forecasting accuracy, measured by APE, under the assumption of comparable levels of actual sales. The premise was that LLMs, by virtue of their capacity to process extensive datasets and discern complex patterns devoid of human cognitive biases, would provide a more data-driven and objective forecasting approach.

Contrary to this hypothesis, the results of the mixed-effects regression analysis revealed a more nuanced picture (see Table 2 and Figure 2). The performance of the LLMs varied, with ChatGPT3.5 and Llama2 yielding significantly higher APE values compared to human forecasters, indicating inferior forecasting accuracy.

On the other hand, Bard, Bing and ChatGPT4 did not show significant differences in APE compared to human forecasters, which indicates that their forecasting accuracy was not notably worse or better than that of humans. These findings challenge the initial hypothesis, revealing that not all LLMs are inherently better at forecasting than humans and that the efficacy of LLMs can be model-specific. Therefore, while some LLMs may have the theoretical capability to outperform human forecasters, the empirical evidence suggests that this capability may not always translate into better forecasting accuracy in practice.

Table 2. mixed-effect model results for H1

|  | APE | | |
|---|---|---|---|
|  | Coefficient | Std. Err. | P-value |
| **Fixed effects** | | | |
| Intercept | 14.35*** | 0.54 | 0.000 |
| Actual sales | -0.00*** | 0.00 | 0.000 |
| Bard | 4.89 | 2.85 | 0.086 |
| Bing | 2.73 | 2.69 | 0.311 |
| ChatGPT3.5 | 14.77*** | 2.80 | 0.000 |
| ChatGPT4 | 0.69 | 2.64 | 0.794 |
| Llama2 | 9.18** | 2.92 | 0.002 |
| **Random effects** | | | |
|  | Estimate | Std. Err. | |
| Forecaster | 1.15 | 0.94 | |
| Residual | 138.77 | 3.68 | |
| Observations | 2973 | | |

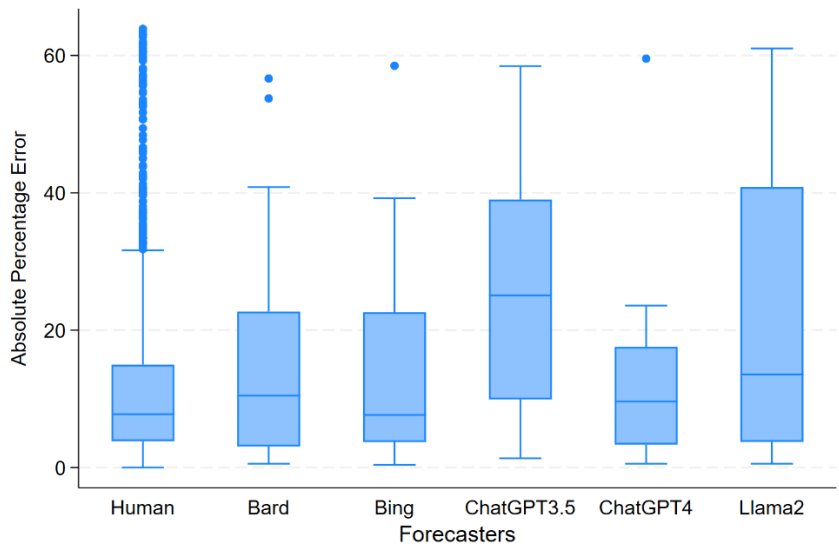

**Figure 2**: APE by Forecaster Type: Human vs. LLMs.*4.2 Enhanced Forecasting Performance with Advanced Forecast Models (H2)*

Considering H2a, which posits that both human forecasters and LLMs would benefit from advanced forecasts, separate models were tested for each forecaster type. In doing so, we conducted a mixed-effects regression analysis for human forecasters to account for within-subject variability and separate regression analyses for each LLM. The results are shown in Appendix J. In the mixed-effects regression analysis, with baseline versus advanced forecasting methods as predictors, advanced forecasts did not significantly reduce accuracy for human forecasters compared to baseline forecasts. Similarly, advanced forecasts did not yield any significant changes in forecast accuracy for each of the LLMs when compared to baseline forecasts. In H2b, we hypothesized that LLMs will show a greater improvement in forecasting performance than human forecasters when using advanced forecasts. The results are shown in Table 3 and Figure 3. Among the different types of forecasters, Bard, ChatGPT3.5, and Llama2 exhibited a significant increase in APE by 10.73%, 10.25%, and 9.79% respectively compared to the performance of human forecasters. This result does not support H2b, as these increases suggest a decline in forecasting performance when these LLMs are provided with advanced forecasts.

However, for Bard, the interaction term with the advanced forecast type significantly decreased APE by -14.64%, partially supporting H2b. This indicates that Bard performed better with advanced forecasts compared to its baseline performance. This interaction effect is noteworthy as it contrasts with the main effect of the advanced forecast type, suggesting that while advanced forecasts did not generally improve performance for forecasters, Bard was able to utilize the advanced information effectively. On the contrary, for ChatGPT3.5 interaction term with the advanced forecast type significantly increased APE by 11.85%, suggesting that ChatGPT3.5 performed worse when provided with advanced forecasts compared to its baseline performance.

Table 3. Mixed-effect model results for H2

|  | APE | | |
| --- | --- | --- | --- |
|  | Coefficient | Std. Err. | P-value |
| **Fixed effects** | | | |
| Intercept | 14.075*** | 0.565 | 0.000 |
| Actual sales | -0.001*** | 0.000 | 0.000 |
| Advanced forecast | 0.807 | 0.455 | 0.076 |
| Bard | 10.727** | 3.572 | 0.003 |
| Bing | 0.665 | 3.334 | 0.842 |
| ChatGPT3.5 | 10.247** | 3.446 | 0.003 |
| ChatGPT4 | 0.047 | 3.234 | 0.988 |
| Llama2 | 9.787** | 3.572 | 0.006 |
| advanced # Bard | -14.643** | 5.386 | 0.007 |
| advanced # Bing | 5.225 | 5.044 | 0.300 |
| advanced # ChatGPT3.5 | 11.846* | 5.303 | 0.025 |
| advanced # ChatGPT4 | 1.704 | 4.978 | 0.732 |
| advanced # Llama2 | -1.645 | 5.610 | 0.769 |
| **Random effects** | | | |
|  | Estimate | Std. Err. | |
| Forecaster | 1.16 | 0.93 | |
| Residual | 138.24 | 3.67 | |
| Observations | 2973 | | |

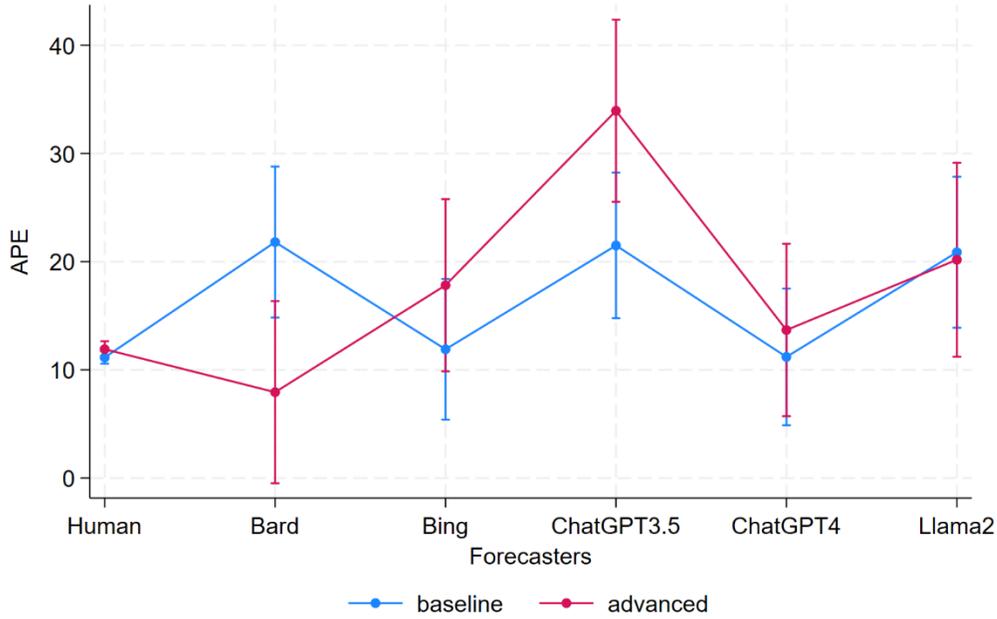

**Figure 3**: Interaction: forecasters and statistical forecast type (Note: the results are predictive margins with 95% CI)

*4.3 Variability in Forecast Accuracy During Promotional and Non-Promotional Periods (H3)*

To investigate the variability in APE across promotional and non-promotional periods for different types of forecasters (H3a), mixed-effects regression analysis for human forecasters and separate regression analyses for each LLM were conducted. The results are shown in Appendix J. The analysis results demonstrate that during promotional periods, human forecasters experienced a significant increase in APE, indicating reduced forecast accuracy. For the LLMs, ChatGPT3.5, and Llama2 showed a notable increase in APE during promotional periods, indicating a decline in forecasting performance compared to non-promotional periods. Conversely, Bard, Bing and ChatGPT4 did not exhibit significant changes in APE during promotional periods.

Regarding H3b, the mixed-effects regression analysis indicated a significant overall increase in APE during promotional periods, implying reduced forecasting accuracy during these times. The results are shown in Table 4 and Figure 4, respectively. Contrary to H3b, which predicted that LLMs would show a smaller difference in APE compared to humans, certain LLMs (ChatGPT3.5, and Llama2) experienced a significant increase in APE during promotional periods. This result suggests a decrease in their forecasting performance during promotional periods, contrary to the hypothesis.

The absence of significant main effects among different forecaster types, with coefficients ranging from -3.10 for ChatGPT3.5 to 0.56 for ChatGPT4, indicates that no single forecaster type consistently outperformed or underperformed human forecasters across conditions. Significant random effects in the model pointed to substantial variability (1.42%) in APE among different forecasters. The inclusion of these random effects substantially improved the model's fit, as evidenced by the likelihood ratio test (p < 0.05).

**Table 4.** Mixed-effect models results for H3

|  | APE | | |
|---|---|---|---|
|  | Coefficient | Std. Err. | P-value |
| **Fixed effects** | | | |
| Intercept | 12.771*** | 0.590 | 0.000 |
| Actual sales | -0.001*** | 0.000 | 0.000 |
| Yes promo | 3.295*** | 0.481 | 0.000 |
| Bard | -0.695 | 4.557 | 0.879 |
| Bing | 0.198 | 4.557 | 0.965 |
| ChatGPT3.5 | -3.100 | 4.557 | 0.496 |
| ChatGPT4 | 0.561 | 4.557 | 0.902 |
| Llama2 | -2.061 | 4.557 | 0.651 |
| yes promo # Bard | 8.871 | 5.452 | 0.104 |
| yes promo # Bing | 3.632 | 5.272 | 0.491 |
| yes promo # ChatGPT3.5 | 26.950*** | 5.384 | 0.000 |
| yes promo # ChatGPT4 | 0.136 | 5.225 | 0.979 |
| yes promo # Llama2 | 18.149** | 5.531 | 0.001 |
| **Random effects** | | | |
|  | Estimate | Std. Err. | |
| Forecaster | 1.42 | 0.94 | |
| Residual | 134.25 | 3.57 | |
| Observations | 2973 | | |

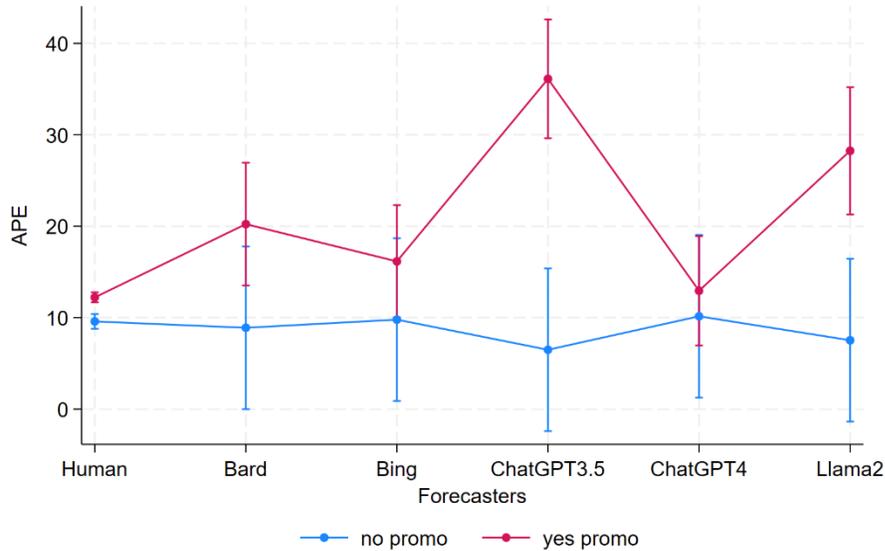

**Figure 4**: Interaction: forecasters and promotion

*4.4 Impact of External Conditions on Forecasting Disparities (H4)*

When addressing H4a, the mixed-effects analysis for human forecasters and regression analyses for each of LLMs were conducted in order to test the forecaster-specific APE under negative vs positive external impact conditions. The results are shown in Appendix J. The parameters indicated substantial variability in forecasting performance among different forecasters. The results suggest that the presence of positive vs negative external impacts on forecasting accuracy did not significantly vary by different forecaster types.

Our examination of the impact of negative versus positive external impacts on APE when comparing the forecast accuracy for LLMs versus human forecasters (H4b) yielded insights that partially counter H4b as shown in Table 5 and Figure 5. Notably, the presence of a positive external impact was not found to be associated with a significant increase in APE. Contrary to the hypothesis, which anticipated a smaller difference in APE for LLMs, ChatGPT3.5 and Llama2 demonstrated significant increases in APE compared to human forecasters, while Bard, Bing and ChatGPT4 did not exhibit significant changes. The interaction effects further suggested a complex response pattern, particularly for Llama2, with a decreased APE (i.e. improved accuracy) under positive impact. These results suggest that the LLMs' algorithmic nature did not uniformly mitigate the impact of positive external influences, as was hypothesized.

Table 5. Mixed-effect model results for H4

|  | APE | | |
|---|---|---|---|
|  | Coefficient | Std. Err. | P-value |
| **Fixed effects** | | | |
| Intercept | 14.417*** | 0.736 | 0.000 |
| Actual sales | -0.001*** | 0.000 | 0.000 |
| Positive (env. impact) | -0.090 | 0.505 | 0.859 |
| Bard | 4.934 | 3.578 | 0.168 |
| Bing | 0.654 | 3.578 | 0.855 |
| ChatGPT3.5 | 14.523*** | 3.578 | 0.000 |
| ChatGPT4 | -3.158 | 3.452 | 0.360 |
| Llama2 | 13.861*** | 3.722 | 0.000 |
| positive # Bard | -0.110 | 5.393 | 0.984 |
| positive # Bing | 4.340 | 4.936 | 0.379 |
| positive # ChatGPT3.5 | 0.569 | 5.212 | 0.913 |
| positive # ChatGPT4 | 8.398 | 4.845 | 0.083 |
| positive # Llama2 | -11.123* | 5.490 | 0.043 |
| **Random effects** | | | |
|  | Estimate | Std. Err. | |
| Forecaster | 1.15 | 0.94 | |
| Residual | 138.68 | 3.68 | |
| Observations | 2973 | | |

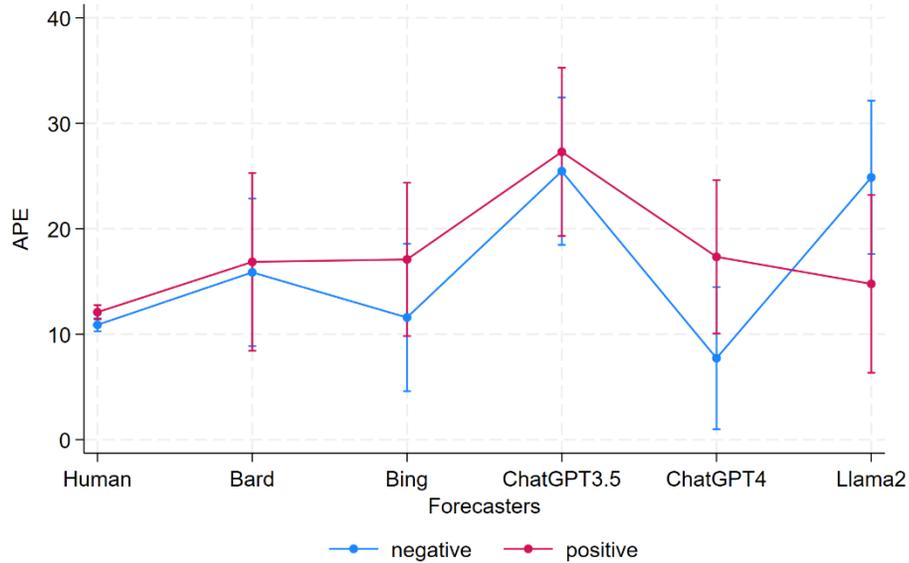

Figure 5: Interaction: Forecasters and external impact

*4.5 Variability in Forecast Accuracy under Complex Forecasting Scenarios (H5)*

In a series of regression analyses comprising mixed-effects model analysis for human forecasters and regression model for each LLM type, we examined the impact of various treatment conditions on APE for each of forecaster types (H5a). In all models, the baseline condition, represented by the scenario "Yes promotion, Positive impact, Baseline statistical forecast," serves as the reference point for comparison. The results are summarized in Appendix J.

For human forecasters, when compared to the baseline, the "Yes promotion, Positive external impact, Advanced forecast" condition significantly increased APE. This suggests that advanced forecasting models, combined with positive external impacts and promotional activities, lead to decreased accuracy in human forecasts. The "Yes promotion, Negative external impact, Baseline forecast" condition also showed an increase in APE compared to the baseline treatment condition, indicating reduced accuracy under this scenario. In contrast, the "No promotion, Positive external impact, Baseline forecast" condition exhibited a significant decrease in APE, indicating an improvement in forecast accuracy compared to the baseline treatment condition. This suggests that human forecasters may better perform when provided with baseline forecasting under positive external impacts during non-promotional periods. This trend indicates that the interplay of promotional conditions, external impacts, and the nature of the forecasting model (baseline vs. advanced) significantly influences the accuracy of human forecasters. Human forecasters' accuracy is likely to improve under non-promotional conditions, particularly in handling positive impacts when they are provided with baseline forecasting method.

Bard showed significant decreases in APE under almost all conditions compared to the baseline treatment condition, except for the condition "Yes promotion, Negative external impact, Baseline forecast." Notably, the most substantial decreases were observed in "Yes promo, Positive external impact, Advanced forecast", "No promo, Positive impact, Baseline forecast", and "No promo, Negative impact, Baseline forecast". This suggests that "Bard" performs better in terms of accuracy in these altered conditions compared to the baseline scenario. The analysis for Bing showed no significant changes in APE across all treatment conditions compared to the baseline treatment condition. The coefficients fluctuated around the baseline without showing a clear trend of increase or decrease, indicating a consistent performance across different scenarios. ChatGPT3.5 demonstrated significant decreases in APE in the "No promo, Positive external impact, Baseline forecast" and "No promo, Negative impact, Baseline forecast" conditions. These results suggest improved accuracy under non-promotional conditions, irrespective of the external impact. Similar to Bing, ChatGPT4 showed no significant change in APE across different treatment conditions compared to the baseline. The coefficients indicate a relatively stable performance across various scenarios. Llama2 exhibited significant decreases in APE in the "Yes promo, Positive impact, Advanced forecast",

"No promo, Positive external impact, Baseline forecast", and "No promo, Negative impact, Baseline forecast" conditions, indicating enhanced forecasting accuracy in these scenarios compared to baseline condition.

In our mixed-effects regression analysis investigating the difference in forecast performance by LLMs vs human forecasters considering the complex nature of the forecasting task as represented by treatment conditions and encompassing the forecast type, promotional conditions, and external impacts (H5b), we observed significant variations in APE across different forecaster types (see Table 6 and Figure 6). The findings suggest that, irrespective of forecaster type, the response to specific treatment conditions across all forecasters varied. In the treatment condition combining promotion with positive external impact and advanced forecasting, overall APE was significantly increased. Similarly, the treatment condition with promotion, negative external impact, and baseline forecasting featured increased APE. In non-promotional period, coupled with positive external impacts and baseline forecasting led to decreased APE.

Regarding main effects of individual forecaster types, significant increases in baseline APE were observed for Bard, ChatGPT3.5, and Llama2 compared to the human forecasters. No significant changes were found for Bing and ChatGPT4 vs human forecasters

Interaction effects between treatment conditions and forecaster types revealed notable variations as shown in Table 6 and Figure 6: Bard and Llama2 exhibited significant reductions in APE in several treatment conditions, particularly under advanced forecasting scenarios. Bing and ChatGPT4 displayed less variation in their performance across the different treatments. Notably, forecast accuracies for ChatGPT3.5 and Llama2 were higher in non-promotional conditions under both negative and positive external impact. Generally, the findings suggest that promotional conditions typically increased APE, with some LLMs showing a capacity to adapt more effectively to these conditions.

**Table 6**. Mixed-effect model results for H5

|  | APE | | |
| --- | --- | --- | --- |
|  | Coefficient | Std. Err. | P-value |
| **Fixed effects** | | | |
| Intercept | 13.170*** | 0.791 | 0.000 |
| Actual sales | -0.001*** | 0.000 | 0.000 |
| Yes promo, Positive impact, Advanced forecast | 4.698*** | 0.784 | 0.000 |
| Yes promo, Negative impact, Baseline forecast | 4.538*** | 0.745 | 0.000 |
| Yes promo, Negative impact, Advanced forecast | -1.054 | 0.718 | 0.142 |
| No promo, Positive impact, Baseline forecast | -2.543** | 0.814 | 0.002 |
| No promo, Negative impact, Baseline forecast | -0.362 | 0.731 | 0.620 |
| Bard | 28.480*** | 8.126 | 0.000 |
| Bing | 7.441 | 5.832 | 0.202 |
| ChatGPT3.5 | 33.223*** | 6.684 | 0.000 |
| ChatGPT4 | 6.817 | 5.832 | 0.242 |
| Llama2 | 28.480*** | 8.126 | 0.000 |
| Yes promo, Positive impact, Advanced forecast # Bard | -36.267*** | 10.355 | 0.000 |
| Yes promo, Positive impact, Advanced forecast # Bing | -2.916 | 8.033 | 0.717 |
| Yes promo, Positive impact, Advanced forecast # ChatGPT3.5 | -14.143 | 9.266 | 0.127 |
| Yes promo, Positive impact, Advanced forecast # ChatGPT4 | -0.932 | 8.033 | 0.908 |
| Yes promo, Positive impact, Advanced forecast # Llama2 | -37.479*** | 10.355 | 0.000 |
| Yes promo, Negative impact, Baseline forecast # Bard | -2.160 | 10.354 | 0.835 |
| Yes promo, Negative impact, Baseline forecast # Bing | -13.483 | 8.671 | 0.120 |
| Yes promo, Negative impact, Baseline forecast # ChatGPT3.5 | -13.759 | 9.265 | 0.138 |
| Yes promo, Negative impact, Baseline forecast # ChatGPT4 | -14.536 | 8.032 | 0.070 |
| Yes promo, Negative impact, Baseline forecast # Llama2 | -2.731 | 10.354 | 0.792 |
| Yes promo, Negative impact, Advanced forecast # Bard | -29.640** | 9.491 | 0.002 |
| Yes promo, Negative impact, Advanced forecast # Bing | -0.674 | 7.620 | 0.930 |
| Yes promo, Negative impact, Advanced forecast # ChatGPT3.5 | -8.887 | 8.290 | 0.284 |
| Yes promo, Negative impact, Advanced forecast # ChatGPT4 | -8.589 | 7.620 | 0.260 |
| Yes promo, Negative impact, Advanced forecast # Llama2 | -7.512 | 9.823 | 0.444 |
| No promo, Positive impact, Baseline forecast # Bard | -26.573* | 10.357 | 0.010 |
| No promo, Positive impact, Baseline forecast # Bing | -5.483 | 8.675 | 0.527 |
| No promo, Positive impact, Baseline forecast # ChatGPT3.5 | -39.651*** | 9.269 | 0.000 |
| No promo, Positive impact, Baseline forecast # ChatGPT4 | -4.941 | 8.675 | 0.569 |
| No promo, Positive impact, Baseline forecast # Llama2 | -30.924** | 10.357 | 0.003 |
| No promo, Negative impact, Baseline forecast # Bard | -31.129** | 9.825 | 0.002 |
| No promo, Negative impact, Baseline forecast # Bing | -8.565 | 8.032 | 0.286 |
| No promo, Negative impact, Baseline forecast # ChatGPT3.5 | -33.828*** | 8.670 | 0.000 |
| No promo, Negative impact, Baseline forecast # ChatGPT4 | -7.244 | 8.032 | 0.367 |
| No promo, Negative impact, Baseline forecast # Llama2 | -30.256** | 9.825 | 0.002 |
| **Random effects** | | | |
| | Estimate | Std. Err. | |
| Forecaster | 1.74 | 0.95 | |
| Residual | 127.95 | 3.41 | |
| Observations | 2973 | | |

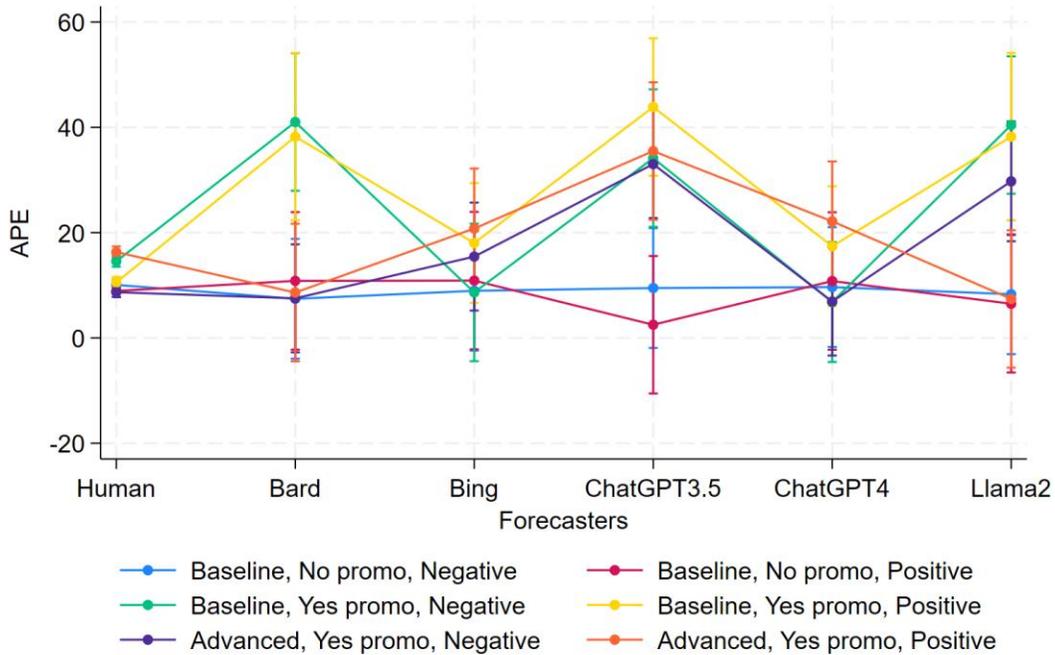

**Figure 6**: Interaction: forecasters and treatment conditions

*4.5 Post-Experimental Questionnaire: Human Forecasters Perspective*

The post-experimental questionnaire was designed to assess the perceived importance of various information types when human participants made judgments using statistical forecasts and to evaluate their agreement with statements about the impact of promotions on their judgment. The results in Appendix F reflect a comparative assessment between baseline and advanced statistical forecasts. Participants rated the importance of information such as forthcoming promotion type, external impact, and historical sales data, with means ranging from 5.11 to 5.68, indicating a moderate to high level of importance across all factors when using both baseline and advanced forecasts. Notably, 'Advanced statistical forecast' was deemed most critical with a mean importance rating of 5.82 when considering advanced forecasts vs baseline forecast (mean importance rating of 5.30). Respondents' agreement with the impact of promotions on their judgment produced mean scores around the mid-5 range, indicating a mild to moderate agreement that both earlier and later promotions influenced their judgment about the forecast in period 25. The self-assessment of the accuracy of participants judgmental forecasts revealed a mean of 4.44 out of 7 (where 7 is 'Very high'), indicating a modestly positive self-assessment of the participants' judgmental forecasting accuracy.

**5. Discussion and conclusion**

This study embarked on an in-depth exploration to compare the forecasting abilities of human experts and LLMs in the retail sector. An experimental setup was devised involving 123 human forecasters, primarily business school graduate students, and five leading LLMs, including Chat GPT3.5, Bard, Llama2, Bing, and ChatGPT-4, to investigate their accuracy in predicting sales data under various conditions such as during normal and promotional sales periods. These LLMs, renowned for their advanced natural language processing capabilities, were evaluated against human judgment, known for its understanding and adaptability in complex forecasting environments. The experiment was structured to cover diverse scenarios, encompassing both baseline and advanced forecasting models, under varying conditions like promotions and external impacts.

The results portrayed a landscape where the effectiveness of LLMs in sales forecasting was not unequivocally superior to that of human forecasters. While some LLMs, such as ChatGPT-4 and Bing, displayed forecasting accuracies on par with human participants, others like ChatGPT3.5 and Llama2 lagged behind. This outcome challenges the prevalent notion that LLMs inherently surpass human forecasters in accuracy. It stresses the importance of considering the model-specific capabilities and the context-dependent nature of these AI tools when deployed in complex forecasting scenarios. The study's findings suggest that while LLMs have considerable potential, their effectiveness is influenced by the nature of the forecasting task and the specific model's characteristics, thus requiring careful consideration when integrating them into practical forecasting processes.

*5.1 Discussion of main findings*

Our hypothesis that LLMs would generally surpass human forecasters in accuracy was not uniformly supported. While some LLM models like ChatGPT4 were found to be generally more accurate in the forecasting experiment, its performance was not certainly consistently better than other LLMs across various conditions. This finding aligns with mixed results in LLM performance reported by Schoenegger and Park (2023) and reflects Makridakis et al. (2023) who caution against assuming uniform superiority of LLMs in forecasting. This study results not only show variations in LLM performance in forecasting but also suggest that the processing capabilities of the LLMs, like those of humans, are different and may vary from one model to another. The varied APE scores of LLMs highlight the importance of a strategic approach in using appropriate models for relevant forecasting tasks.

The expectation that using advanced forecasting models would universally enhance human forecasting performance was not confirmed, challenging the assumption that more sophisticated models invariably improve predictions. This outcome is consistent with findings from Fildes & Goodwin (2007) and highlights the intricacies of advanced models, which may sometimes misalign with both human and AI forecasting capabilities as discussed by Kourentzes & Petropoulos (2016). Furthermore, our findings indicate that both human and AI forecasters, including LLMs like Bard, ChatGPT3.5, and Llama2, struggle with maintaining consistent accuracy across promotional and non-promotional periods. This is particularly evident in dynamic environments where unpredictability and volatility challenge the integration of relevant information into forecasts, as noted by Lawrence et al. (2006) and Fildes & Hastings (1994). These observations suggest a need for more adaptive and intuitive forecasting systems that can better handle the complexities of such scenarios, as suggested by Sroginis et al. (2023).

The significant reduction in APE for Llama2 under positive external impact conditions partially supports our hypothesis that LLMs will exhibit varying levels of accuracy depending on the nature of external impacts—positive or negative. Although other LLMs and humans demonstrated similar accuracy under both positive and negative external impact conditions, their response patterns diverged markedly from that of Llama2. Unlike Llama2, which made minimal to no adjustments under positive conditions but significant negative adjustments under adverse conditions, forecasters such as humans, Bing, and ChatGPT responded differently. These forecasters made substantial positive adjustments in response to positive external impacts and comparatively smaller adjustments under negative conditions, as detailed in Appendix E. This varied response strategy resulted in a consistent level of accuracy across both conditions. This finding aligns with the research by Fildes and colleagues (2009), which suggests that larger adjustments typically lead to more significant improvements in accuracy. This is particularly evident in the case of human forecasters, whose average accuracy under these conditions was relatively better than that of ChatGPT3.5 and Llama2. Specifically, in the case of ChatGPT, which made large negative adjustments under negative external impacts and relatively small positive adjustments under positive conditions, the pattern of response underscores the complexities of forecasting accuracy across different contexts.

Looking at trends in forecast accuracy in terms of general descriptive statistics of Median MdAPE and Mean MdAPE (Appendix D), we observe humans generally achieve the lowest median and mean MdAPE across all categories of conditions, suggesting a higher overall accuracy compared to the other forecasters. ChatGPT3.5, on the other hand, shows remarkably higher errors, particularly under promotional and advanced forecasting conditions. In contrast, Bard and Bing improve substantially, with particularly low errors for Bard in advanced forecast conditions, suggesting its strength in more complex scenarios.

We also looked at the average forecast accuracy of various forecasters including experiment participants, experts, and LLMs, as well as their adjustment size to provided forecasts. The results are summarized in Appendices C, D, and E. Looking at trends in forecast accuracy in terms of general descriptive statistics of Median MdAPE and Mean MdAPE (Appendix D), we observe humans generally achieve the lowest median and mean MdAPE across all categories of conditions, suggesting a higher overall accuracy compared to the other forecasters. This is in line with the findings by Schoenegger et al., (2024a), who showed that the collective wisdom of crowd outperforms LLMs in various decision-making scenarios. Additionally, forecast errors tend to increase for almost all forecasters under promotional conditions, with ChatGPT3.5 showing the largest increase. In contrast, humans and ChatGPT4 maintain relatively lower errors, effectively managing the complexities associated with promotional effects. In the absence of promotions, most models demonstrate decreased mean and median MdAPE rates, with ChatGPT4 and Llama2 notably excelling, indicating better performance in stable market conditions. Notably, under advanced forecasting conditions, Bard significantly improves, showing the lowest errors, which suggests that some models may have specialized capabilities that are context dependent. Under positive external impacts, most models except humans tend to struggle with increased errors, highlighting a general sensitivity to favorable external stimuli.

Looking at the adjustment sizes to forecasts (Appendix E), distinct trends can be observed among different forecasters. Human forecasters typically exhibit the largest mean and median adjustments, indicating a robust and adaptive response to changing market conditions, particularly during promotional periods and under both positive and negative external impacts. In contrast, models like ChatGPT3.5 and Llama2 often show reduced mean and median adjustments, particularly in complex and adverse conditions like advanced forecasting, promotional periods, and negative external impacts, suggesting a conservative approach. Our observations reveal that such smaller adjustments often correlate with increased error rates. This is particularly notable in models like Bard during advanced forecasting conditions where its minimal adjustment of -39.4 corresponded with poorer performance metrics. This trend suggests that overly conservative adjustments might not adequately address the complexities of certain forecasting scenarios, thereby potentially compromising accuracy. This perspective is supported by Fildes et al. (2009), who argued that minor adjustments generally do not enhance forecast accuracy, and it may sometimes be preferable to maintain the original forecast in the face of minor changes. Bing and ChatGPT4 display variability, with Bing making large adjustments under positive external impacts but decreasing adjustments under advanced forecast conditions. ChatGPT4, similarly, adapts well under promotional and positive conditions, showing a capacity for considerable adjustments akin to human forecasters. This is in line with the findings by Fildes at al. (2009) who found large adjustments by human experts are often useful to improve the forecast accuracy. Our results show that positive adjustment is associated with lower forecast error on average. However, this may not hold for different conditions. We have a limited number of periods with no adjustment, and we observe positive adjustments more frequently than negative adjustments. This might be because we have had many promotion periods in series and naturally forecasts needed to be adjusted positively. For details, please see Appendix E.

*5.2 Implications*

This research significantly contributes to the body of forecasting literature by empirically examining and comparing the efficacy of human forecasters and LLMs in forecasting sales under different conditions. It enhances our understanding of the complex relationship between human judgment and AI in forecasting, and challenges the assumption that LLMs are inherently superior to human forecasters. The study's findings, particularly regarding the differential performance of LLMs and the complex interaction of external factors in forecasting accuracy, provide a deeper understanding of the capabilities and limitations of both human and AI forecasters. This research highlights the importance of context and model specificity in forecasting, aligning with the theoretical frameworks suggested by Fildes & Goodwin (2007) and Lawrence et al. (2006), and offers a refined perspective on the integration of human expertise with advanced AI models in forecasting.

The study's insights are particularly valuable for practitioners in the retail industry, emphasizing the importance of a carefully considered approach when integrating LLMs into forecasting processes. Given the variability in LLM performance, especially during complex promotional periods as highlighted in our findings, it becomes imperative for industry professionals to not overly rely on these models without due consideration of their context-specific capabilities and limitations. This research advocates for a strategic blend of AI and human judgment, tailored to the nuances of specific forecasting scenarios. The findings suggest that over-reliance on LLMs without accounting for their limitations in certain contexts, particularly during promotional periods as indicated by the increased APE values, may lead to suboptimal forecasting outcomes. Therefore, retail practitioners should consider these insights when designing and implementing forecasting systems, ensuring they are equipped to handle the intricacies of various sales contexts effectively.

The implications of this study extend to policymakers and industry regulators, who should consider these findings when establishing guidelines and frameworks for the adoption and integration of AI in business forecasting practices. The study highlights the necessity of policy frameworks that acknowledge the limitations and strengths of both human forecasters and LLMs. Regulatory guidelines should encourage the development of AI systems that are not only advanced in terms of data processing but also capable of being effectively integrated with human judgment. This is particularly relevant in the context of the increasing reliance on AI for decision-making in various industries.

*5.3 Limitations*

A notable limitation of this study is the reliance on graduate students as the primary human forecasters. While these participants possess academic knowledge and some practical skills in business forecasting, they may not have the extensive practical experience and in-depth understanding typical of seasoned industry professionals. This gap potentially affects the generalizability of the findings, as real-world business forecasting often involves intricate decision-making and judgment calls that are honed through extensive professional experience. The diversity in expertise and approach that experienced professionals bring to forecasting could yield different insights, especially in terms of how they interact with and interpret the outputs of advanced models like LLMs. Hence, while the study provides valuable insights into the comparative abilities of novice human forecasters and LLMs, it may not fully reflect the forecasting landscape in a professional setting.

The other limitation of this study relates to the specific data used in this study. We provided 24 time series from real-world data for FMCG products that represent different behaviors in trend, promotion, etc. However, it will be useful to replicate this study on another dataset with different patterns and from industries to evaluate whether the results hold or not. While we looked at only two variables "Promotion Status" and "External Impact" variables, other variables with different settings can be considered. For instance, "External Impact" in our experiment is considered a binary variable with "Positive" and "Negative" impact such that they can alter sales up to 10%, We were inspired by the case study in our experimental design, however, this variable can take different forms and have different impact on sales and might be considered in future research. The study also reveals a limitation in the generalizability of its findings across various AI models due to the variable performance of different LLMs. This variability necessitates a more in-depth exploration into the specific traits and training that contribute to the forecasting efficacy of these models. The LLMs selected for the study, including ChatGPT4, represent only a fraction of the rapidly evolving landscape of AI forecasting tools. Future advancements in AI technology and new model developments could significantly alter the efficacy landscape, leading to different outcomes than those observed in the current study. This limitation emphasizes the need for continuous evaluation of AI models in forecasting, taking into account the evolving nature of AI capabilities and the specific characteristics of each model that may impact its performance in various forecasting scenarios.

While LLMs have shown remarkable capabilities in generating human-like text and understanding context, their application in forecasting presents specific challenges, one of the most notable being the tendency for "hallucinations," or the generation of plausible but factually incorrect information. This phenomenon can be originated from the models' reliance on the provided training data, without an inherent understanding of truth or the ability to verify facts against real-world developments (Bender et al., 2021; Ribeiro et al., 2020). Consequently, while LLMs can extrapolate trends and patterns, their predictive outputs may be compromised by inaccuracies or fabrications that may seem consistent but disconnected from reality. Moreover, the absence of real-time data integration further limits their forecasting reliability, as they cannot accommodate information after their training. These limitations require careful consideration and additional validation when using LLMs for forecasting purposes, highlighting the need for human oversight and the incorporation of current, domain-specific data.

*5.4 Future research*

The limitations identified call for future research that incorporates a broader spectrum of human forecasters, particularly those with extensive professional experience in the retail sector. Engaging experienced industry professionals would enhance the external validity of the findings and provide insights into how seasoned forecasters interact with and interpret the outputs of AI models like LLMs. Additionally, future studies should explore a wider range of forecasting scenarios, including those that mimic real-world complexities and uncertainties more closely. This approach would help to understand how both human forecasters and LLMs perform under various market conditions, such as fluctuating demand, unexpected market shifts, and significant events. By examining a more diverse set of scenarios, researchers can gain a deeper understanding of the strengths and weaknesses of both human and AI forecasters in different contexts.

Further research is essential to explore how various training datasets, model architectures, and contextual inputs influence the forecasting accuracy of different LLMs, especially in complex retail environments. Investigating these aspects could reveal critical insights into the specific characteristics that enhance the forecasting capabilities of LLMs. Future studies should investigate how these two approaches can complement each other most effectively, potentially through the development of hybrid forecasting models. Such models could dynamically adjust the contribution of human and AI elements based on specific conditions, optimizing accuracy and reliability in diverse forecasting scenarios. Building on the exploratory findings from Schoenegger et al. (2024a), which highlighted the potential of LLMs to significantly enhance human forecasting accuracy even when models are intentionally biased, future research should delve deeper into the mechanisms by which LLMs influence human judgment and decision-making in forecasting tasks. This research could explore the balance between AI's data-processing capabilities and human intuition and expertise, aiming to develop forecasting systems that leverage the strengths of both while mitigating their limitations.

To further extend the findings of this study, future research should explore a wider array of LLMs, including emerging models with potentially different capabilities and limitations. Alongside this, the application of advanced analytical techniques, such as machine learning algorithms capable of handling non-linear relationships and high-dimensional data, could be explored. In our experiment, we trained the LLMs with relatively small data that are specific to the case study, although the LLM models are trained offline on a massive amount of data. Future studies might consider developing specific LLMs with in-house data and rules to mimic human behavior and reasoning. Such a system can be used as a decision support systems where experts keep a log of their data including their adjustment, accuracy performance, and reasoning for adjustments, and call upon them in future when required. There is also a potential for developing prompt-based forecasting models that are able to build models on companies' in-house data and provide forecasts. Although this is a challenging task, the practice for simple models already exists. This approach would provide further understanding of how different AI models perform across a variety of forecasting tasks, offering insights into the evolution of AI in the field of forecasting. It will also contribute to the ongoing discourse on the integration of AI and human expertise in developing advanced forecasting systems and strategies.

Finally, future studies should explore different configurations of LLM interaction - from passive to highly interactive - to determine their effects on the range and variance of predictions. This is crucial for understanding whether LLMs might undermine the wisdom-of-the-crowd effect, which is vital in many forecasting contexts (Schoenegger et al., 2024b). Further research should also consider the long-term effects of repeated LLM use on human forecasting skills. Investigating whether reliance on LLMs leads to skill enhancement or degradation over time could inform how best to integrate LLMs in practice without diminishing human expertise (Schoenegger et al., 2024a).

**5.5 Concluding remarks**

This study's examination of the forecasting abilities of both human experts and LLMs in retail sales forecasting provided several key insights. The experiments, which involved human forecasters, primarily business school graduate students, and five leading LLMs including ChatGPT4, revealed that the performance of LLMs in forecasting is not uniformly superior to that of humans. While some models like ChatGPT4 and Bing demonstrated forecasting accuracies comparable to humans, others such as ChatGPT3.5 and Llama2 did not perform as well. These findings suggest that the effectiveness of LLMs in forecasting tasks varies depending on the specific model and the nature of the task at hand.

The study highlighted that advanced forecasting models do not automatically enhance forecasting performance. Both human forecasters and LLMs displayed limitations when working with these models, indicating that the integration of sophisticated models into forecasting requires careful consideration. This aligns with previous research suggesting that the complexity of advanced models can sometimes outweigh their potential benefits in accuracy. Furthermore, the research revealed that both human and AI forecasters struggle with the unpredictability of promotional periods, as evidenced by increased APE values during these times. This suggests a shared challenge in adapting to dynamic market conditions and underscores the need for forecasting models that can better handle such complexities. This study also points to the importance of further research to explore the integration of human judgment and advanced AI models in forecasting to optimize their combined strengths.

**Appendix A: Experiment guidelines to human forecasters**

Dear participant,

We would like to thank you for taking part in this experiment. The current experiment is designed to test a few key concepts in retail sales forecasting. All results and information are confidential and will be only used for research purposes. Please note, by taking part and submitting your results, you agree with the conditions of the study outlined below.

You will be given 24 different sales series representing the actual sales of different products in a company. Each series has 24 observations, or periods, and you are asked to act as the sales manager and forecast sale at period 25 based on the given information. We assume that all of the internal and external variables that contribute to sales are summarized to the given information. Your performance will be evaluated based on the accuracy of your forecast. Mean Absolute Percentage Error (MAPE) of the forecasts (which will be calculated automatically) is considered as the accuracy criteria. Based on the aggregated accuracy of your forecasts, at the end of the experiment you will receive a monetary reward. The higher the accuracy of your forecast, the higher is the reward that you will receive. The reward will not be higher than 20 AUD and not smaller than 5 AUD.

Figure 1 shows a screenshot of the experiment interface. It represents the dashboard of the forecasting system that you will use to support your forecasting decisions.

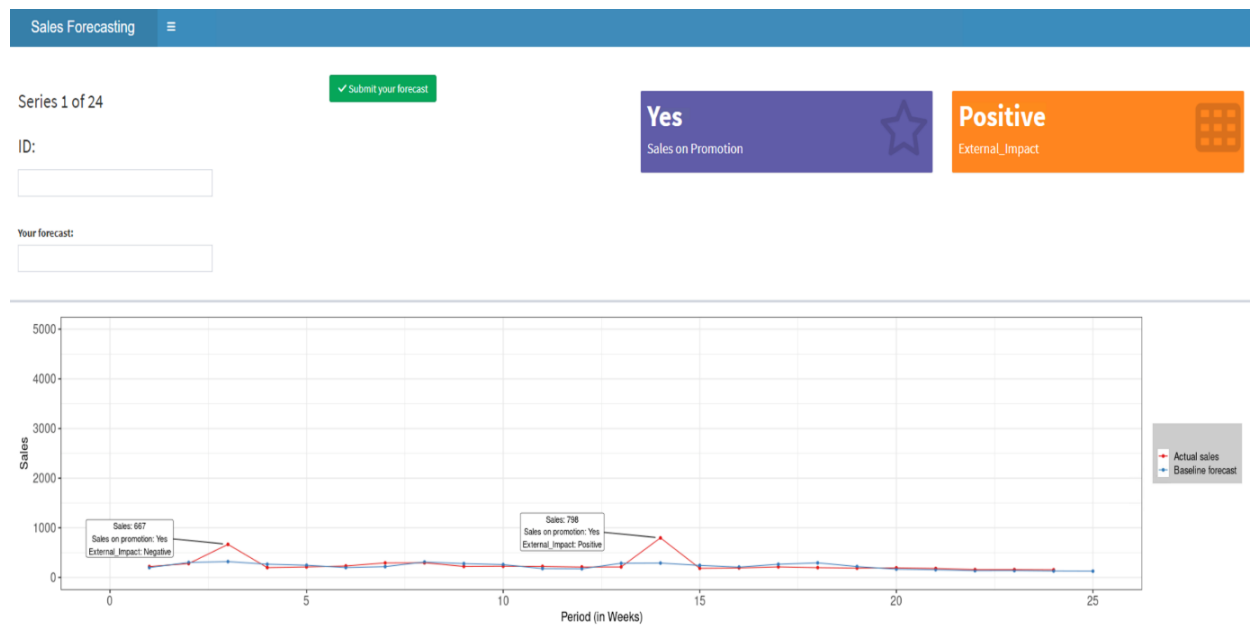

**Figure 1:** Interface of the experiment

**Provided Information:**

The given information includes time series observations (with relevant information such as promotion, and external impact for some existing time points), promotion type and external impact at period 25, and system statistical forecast (Baseline or advanced statistical forecast for all periods).

"**Promotion type**" is one of the important variables that causes sales uplifts. The variable "Promotion" status can be Yes and No Pro, meaning that promotion exist and do not exist at the corresponding period. The size of promotion uplifts varies depending on the product as it is evident in the corresponding time series.

The other variable is **"External impact"** which effectively summarises competitors' activity and whether that has impacted/ or will impact our sales "positively" or "negatively". This might be different during promotional and non-promotional periods. **"External impact"** status can be either "Positive" or "Negative".

Beside this information, you will be given a "**statistical forecast**". The statistical forecast may be of two types: "**baseline forecast**" or "**advanced forecast**". The "baseline statistical forecast" is the forecast generated by a software in the company, which uses "simple exponential smoothing" model and **does not consider the possible impact of "promotion"** and **"external impact"** on sales. Whereas the **"advanced forecast"** takes into account the promotion impact and uses a "dynamic linear regression" model to forecast sales.

**Forecast needed:**

You are asked to sit on the corresponding experiment based on your ID and then enter your ID, provided to you by the research facilitator at the beginning of the session. Please note, the failure to enter will not allow us to pay for your forecasting accuracy at the end of the experiment. You will then have to predict the value of the sales at period 25 for all given series and enter (type in) your forecast in the white text box "Your forecast: "on the top left of the screen and submit your forecast by pressing the green button "submit your forecast".  Then, you can click on the "Next series" and move to the next time series. Please note that once you have submitted your forecast, you cannot go back and edit your forecast. There is no time limit and you can think about each forecast as much as you want. **At the end, you are required to save the results by clicking on "save results".** Once you save the results, a message will pop up as shown below.

> **Results saved!**
>
> Thank you for your participation, your results have been saved. You can now close the webpage.
>
> [Dismiss]

You can close the webpage and let the coordinator know.

**We would like to remind you again: please sit at your experiment as indicated with ID** as you will be given payment between A$ 5 and A$ 20 based on the accuracy of your forecasts.

Thank you again for participating and contributing to this research.

Happy forecasting,

The Research Team

**Appendix B: LLM prompts.**

As mentioned before, in our prompt we tried to provide the same information to LLM as it was given to participants. This includes using the same wording and keeping the same sequence of information according to the instruction given to students. We removed the irrelevant instructions from the handout. The information was given to LLM in table format, rather than time series plot as it was given to participants. The given prompt for all LLMs is as follows:

We are going to run an experiment. The current experiment is designed to test a few key concepts in retail sales forecasting. All results and information are confidential and will be only used for research purposes. You will be given 24 different sales series representing the actual sales of different products in a company. Each series has 24 observations, or periods, and you are asked to act as the sales manager and forecast sale at period 25 based on the given information. We assume that all of the internal and external variables that contribute to sales are summarized to the given information. Your performance will be evaluated based on the accuracy of your forecast. Mean Absolute Percentage Error (MAPE) of the forecasts (which will be calculated automatically) is considered as the accuracy criteria. The given information includes time series observations (with relevant information such as promotion, and external impact for some existing time points), promotion type denoted as promo and external impact at period 25, and system statistical forecast (Baseline or advanced statistical forecast for all periods). You will be given the information in the following format: {time, actual_sale, baseline_forecast, advanced_forecast, promo, External_Impact} "Promotion type" is one of the important variables that causes sales uplifts. The variable "promo" status can be Yes and No Pro, meaning that promotion exist and do not exist at the corresponding period. The size of promotion uplifts varies depending on the product as it is evident in the corresponding time series. The other variable is "External-Impact" which effectively summarises competitors' activity and whether that has impacted/ or will impact our sales "positively" or "negatively". This might be different during promotional and non-promotional periods. "External-Impact" status can be either "Positive" or "Negative". Beside this information, you will be given a "statistical forecast". The statistical forecast may be of two types: "baseline forecast" or "advanced forecast". The "baseline forecast" is the forecast generated by a software in the company, which uses "simple exponential smoothing" model and does not consider the possible impact of "promotion" and "external-impact" on sales. Whereas the "advanced forecast" takes into account the promotion impact and uses a "dynamic linear regression" model to forecast sales. Given the provided information, predict the value of the sales in period 25.

**APPENDIX C: Descriptive Statistics of experiment data**

| Conditions | Count | Mean | Median | Min | Max | SD |
|---|---|---|---|---|---|---|
| Overall | 600 | 1681 | 1507 | 80 | 4981 | 1164 |
| Baseline forecast | 375 | 1777 | 1572 | 127 | 4324 | 1159 |
| Advanced forecast | 225 | 1408 | 1210 | 90 | 4601 | 1018 |
| No promotion | 535 | 1581 | 1425 | 80 | 4324 | 1088 |
| Yes promotion | 65 | 2399 | 2324 | 127 | 4981 | 1364 |
| Positive external impact | 40 | 2228 | 2180 | 127 | 4981 | 1273 |
| Negative external impact | 25 | 1547 | 1386 | 80 | 4272 | 1072 |

**APPENDIX D: - Forecasting accuracy of forecasters in terms of APE**

|  | Median MdAPE | Mean MdAPE |
|---|---|---|
| **Overall** | | |
| Human | 8.07 | 8.39 |
| Bard | 10.46 | 10.46 |
| Bing | 7.66 | 7.66 |
| ChatGPT3.5 | 25.08 | 25.08 |
| ChatGPT4 | 9.63 | 9.63 |
| Llama2 | 13.54 | 13.54 |
| **Baseline forecast** | | |
| Human | 7.47 | 8.67 |
| Bard | 17.49 | 17.49 |
| Bing | 8.46 | 8.46 |
| ChatGPT3.5 | 18.19 | 18.19 |
| ChatGPT4 | 8.95 | 8.95 |
| Llama2 | 15.24 | 15.24 |
| **Advanced forecast** | | |
| Human | 7.6 | 7.86 |
| Bard | 3.89 | 3.89 |
| Bing | 7.66 | 7.66 |
| ChatGPT3.5 | 31.88 | 31.88 |
| ChatGPT4 | 10.3 | 10.3 |
| Llama2 | 5.3 | 5.3 |
| **Yes promotion** | | |
| Human | 7.96 | 8.75 |
| Bard | 19.82 | 19.82 |
| Bing | 10.25 | 10.25 |
| ChatGPT3.5 | 35.76 | 35.76 |
| ChatGPT4 | 10.3 | 10.3 |
| Llama2 | 23.28 | 23.28 |
| **No promotion** | | |
| Human | 7.14 | 8.01 |
| Bard | 7.38 | 7.38 |
| Bing | 5.68 | 5.68 |
| ChatGPT3.5 | 4.54 | 4.54 |
| ChatGPT4 | 5.2 | 5.2 |
| Llama2 | 4.99 | 4.99 |
| **Positive external impact** | | |
| Human | 7.44 | 8.46 |
| Bard | 15.79 | 15.79 |
| Bing | 12.84 | 12.84 |
| ChatGPT3.5 | 25.08 | 25.08 |
| ChatGPT4 | 14.48 | 14.48 |
| Llama2 | 9.27 | 9.27 |
| **Negative external impact** | | |
| Human | 8.13 | 8.78 |
| Bard | 6.34 | 6.34 |
| Bing | 7.36 | 7.36 |
| ChatGPT3.5 | 25.03 | 25.03 |
| ChatGPT4 | 4.65 | 4.65 |
| Llama2 | 16.93 | 16.93 |

**APPENDIX E:** Mean and Median of judgmental adjustments by treatment conditions and forecaster types

| | Mean Adjustment size | Median Adjustment size | Mean MdAPE Positive adjustment | Median MdAPE Positive adjustment | Mean MdAPE Negative adjustment | Median MdAPE Negative adjustment | Mean MdAPE No adjustment | Median MdAPE No adjustment |
|---|---|---|---|---|---|---|---|---|
| Overall | | | | | | | | |
| Human | 443 | 129 | 8.3 (n=1930) | 7.8 | 9.1 (n=934) | 8.7 | 12.7 (n=2) | 12.7 |
| Bard | -5.8 | 0 | 1.5 (n=1) | 1.5 | 4.0 (n=3) | 4.0 | 15.2 (n=16) | 15.2 |
| Bing | 149 | 83 | 7.7 (n=15) | 7.7 | 12.9 (n=8) | 12.9 | - | - |
| ChatGPT3.5 | -313 | -135 | 19.0 (n=10) | 19.0 | 30.9 (n=11) | 30.9 | - | - |
| ChatGPT4 | 435 | 125 | 10.3 (n=17) | 10.3 | 13.6 (n=5) | 13.6 | 0.6 (n=2) | 0.6 |
| Llama2 | -131 | 0 | 10.9 (n=5) | 10.9 | 48.9 (n=3) | 48.9 | 5.3 (n=11) | 5.3 |
| Baseline forecast | | | | | | | | |
| Human | 708 | 298 | 8.2 (n=1393) | 7.2 | 11.5 (n=413) | 10.8 | | |
| Bard | 16.6 | 0 | 1.5 (n=1) | 1.5 | 19.8 (n=1) | 19.8 | 17.5 (n=10) | 17.5 |
| Bing | 505 | 347 | 8.5 (n=12) | 8.5 | 12.2 (n=2) | 12.2 | - | - |
| ChatGPT3.5 | 15.4 | 73 | 19.2 (n=9) | 19.2 | 12.6 (n=4) | 12.6 | - | -- |
| ChatGPT4 | 706 | 286 | 9.0 (n=13) | 9.0 | 14.2 (n=2) | 14.2 | - | - |
| Llama2 | 33.8 | 0 | 7.9 (n=4) | 7.9 | 19.8 (n=1) | 19.8 | 16.9 (n=7) | 16.9 |
| Advanced forecast | | | | | | | | |
| Human | -10.5 | 8.5 | 9.3 (n=537) | 8.2 | 7.9 (n=521) | 7.3 | 12.7 (n=2) | 12.7 |
| Bard | -39.4 | 0 | - | - | 3.2 (n=2) | 3.2 | 4.5 (n=6) | 4.5 |
| Bing | -404 | -44 | 7.7 (n=3) | 7.7 | 12.9 (n=6) | 12.9 | - | - |
| ChatGPT3.5 | -846 | -874 | 18.8 (n=1) | 18.8 | 32.6 (n=7) | 32.6 | - | - |
| ChatGPT4 | -16.1 | 0 | 14.2 (n=4) | 14.2 | 13.6 (n=3) | 13.6 | 0.6 (n=2) | 0.6 |
| Llama2 | -414 | 0 | 19.7 (n=1) | 19.7 | 55.0 (n=2) | 55.0 | 2.9 (n=4) | 2.9 |
| Yes promotion | | | | | | | | |
| Human | 624 | 248 | 8.7 (n=1425) | 8.1 | 10.0 (n=579) | 8.8 | 12.7 (n=2) | 12.7 |
| Bard | -24.3 | 0 | - | - | 4.0 (n=3) | 4.0 | 22.7 (n=10) | 22.7 |
| Bing | 222 | 230 | 10.3 (n=10) | 10.3 | 12.9 (n=6) | 12.9 | - | - |
| ChatGPT3.5 | -435 | -219 | 41.7 (n=6) | 41.7 | 31.9 (n=8) | 31.9 | - | - |
| ChatGPT4 | 596 | 219 | 12.4 (n=12) | 12.4 | 13.6 (n=3) | 13.6 | 0.6 (n=2) | 0.6 |
| Llama2 | -238 | 0 | 23.2 (n=2) | 23.2 | 48.9 (n=3) | 48.9 | 5.3 (n=7) | 5.3 |
| No promotion | | | | | | | | |
| Human | 18.3 | 27 | 7.6 (n=505) | 6.9 | 9.1 (n=355) | 9.5 | - | - |
| Bard | 28.6 | 0 | 1.5 (n=1) | 1.5 | - | - | 10.5 (n=6) | 10.5 |
| Bing | -16.6 | 40 | 1.4 (n=5) | 1.4 | 12.2 (n=2) | 12.2 | - | - |
| ChatGPT3.5 | -69.1 | 32 | 2.2 (n=4) | 2.2 | 7.0 (n=3) | 7.0 | - | - |
| ChatGPT4 | 42.7 | 58 | 5.0 (n=5) | 5.0 | 14.2 (n=2) | 14.2 | - | - |
| Llama2 | 52 | 0 | 5.0 (n=3) | 5.0 | - | - | 8.7 (n=4) | 8.7 |
| Positive external impact | | | | | | | | |
| Human | 458 | 180 | 8.4 (n=1031) | 7.7 | 9.9 (n=253) | 7.9 | 12.7 (n=2) | 12.7 |
| Bard | -0.1 | 0 | | | 19.8 (n=1) | 19.8 | 13.5 (n=7) | 13.5 |
| Bing | 402 | 240 | 12.8 (n=9) | 12.8 | 11.5 (n=2) | 11.5 | - | - |
| ChatGPT3.5 | -127 | 46 | 18.8 (n=5) | 18.8 | 32.0 (n=4) | 32.0 | - | - |
| ChatGPT4 | 389 | 148 | 16.3 (n=8) | 16.3 | 14.2 (n=2) | 14.2 | 0.6 (n=1) | 0.6 |
| Llama2 | 19.5 | 0 | 12.3 (n=2) | 12.3 | 19.8 (n=1) | 19.8 | - | - |
| Negative external impact | | | | | | | | |
| Human | 430 | 76 | 8.7 (n=899) | 7.5 | 9.8 (n=681) | 9.8 | - | - |
| Bard | -9.5 | 0 | 1.5 (n=1) | 1.5 | 3.2 (n=2) | 3.2 | 16.9 (n=6) | 16.9 |
| Bing | -82.7 | -18 | 6.4 (n=6) | 6.4 | 12.9 (n=6) | 12.9 | - | - |
| ChatGPT3.5 | -453 | -209 | 19.2 (n=5) | 19.2 | 30.9 (n=7) | 30.9 | - | - |
| ChatGPT4 | 474 | 113 | 4.7 (n=9) | 4.7 | 13.6 (n=3) | 13.6 | 0.6 (n=1) | 0.6 |
| Llama2 | -241 | 0 | 10.9 (n=3) | 10.9 | 55.0 (n=2) | 55.0 | 11.1 (n=6) | 11.1 |

**APPENDIX F:** Forecast accuracy of Expert's in terms of APE (n=1, based on one expert data)

| Expert (n=1) | Mean APE | Median APE |
| --- | --- | --- |
| Overall | 6.12 | 5.64 |
| Baseline forecast | 5.57 | 3.17 |
| Advanced forecast | 7.05 | 6.78 |
| Yes promotion | 6.84 | 6.78 |
| No promotion | 4.39 | 3.17 |
| Positive external impact | 4.63 | 2.11 |
| Negative external impact | 7.38 | 6.03 |

**APPENDIX J:** Mixed-effects model for human forecasters and regression model for each LLM by hypothesis

| | | H2a | | | | | | H3a | | | | | | H4a | | | | | | H5a | | | | | |
|---|---|---|---|---|---|---|---|---|---|---|---|---|---|---|---|---|---|---|---|---|---|---|---|---|---|
| | | Human | Bard | Bing | GPT3.5 | GPT4 | Llama2 | Human | Bard | Bing | GPT3.5 | GPT4 | Llama2 | Human | Bard | Bing | GPT3.5 | GPT4 | Llama2 | Human | Bard | Bing | GPT3.5 | GPT4 | Llama2 |
| | Observations | 2866 | 20 | 23 | 21 | 24 | 19 | 2866 | 20 | 23 | 21 | 24 | 19 | 2866 | 20 | 23 | 21 | 24 | 19 | 2866 | 20 | 23 | 21 | 24 | 19 |
| **Fixed effects** | | | | | | | | | | | | | | | | | | | | | | | | | |
| Intercept | Coef. | 14.112 | 20.467 | 13.700 | 22.216 | 21.234 | 18.021 | 12.734 | 10.608 | 12.150 | 10.991 | 19.702 | 9.307 | 14.525 | 13.545 | 9.505 | 23.638 | 15.219 | 27.447 | 12.965 | 50.404 | 17.665 | 50.876 | 23.813 | 67.008 |
| | Std. err. | 0.564 | 9.738 | 7.902 | 9.616 | 6.573 | 13.303 | 0.590 | 10.135 | 8.410 | 6.568 | 6.890 | 11.579 | 0.735 | 13.116 | 10.458 | 13.470 | 8.448 | 15.602 | 0.794 | 12.734 | 12.306 | 9.818 | 9.788 | 17.799 |
| | P-value | 0.000 | 0.051 | 0.098 | 0.033 | 0.004 | 0.194 | 0.000 | 0.310 | 0.164 | 0.112 | 0.009 | 0.433 | 0.000 | 0.316 | 0.374 | 0.096 | 0.086 | 0.098 | 0.000 | 0.002 | 0.170 | 0.000 | 0.026 | 0.003 |
| Actual sales | Coef. | -0.001 | 0.001 | -0.001 | -0.000 | -0.004 | 0.001 | -0.001 | -0.001 | -0.001 | -0.002 | -0.004 | -0.001 | -0.001 | 0.001 | 0.001 | 0.001 | -0.002 | -0.001 | -0.001 | -0.004 | 0.000 | -0.003 | -0.003 | -0.009 |
| | Std. err. | 0.000 | 0.003 | 0.003 | 0.003 | 0.002 | 0.004 | 0.000 | 0.003 | 0.003 | 0.002 | 0.002 | 0.004 | 0.000 | 0.004 | 0.003 | 0.004 | 0.003 | 0.005 | 0.000 | 0.003 | 0.004 | 0.003 | 0.003 | 0.004 |
| | P-value | 0.000 | 0.876 | 0.794 | 0.931 | 0.094 | 0.810 | 0.000 | 0.828 | 0.708 | 0.368 | 0.073 | 0.845 | 0.000 | 0.849 | 0.829 | 0.884 | 0.344 | 0.858 | 0.000 | 0.239 | 0.968 | 0.323 | 0.394 | 0.059 |
| Advanced forecast | Coef. | 0.806 | -13.895 | 5.992 | 12.500 | 2.573 | -0.565 | | | | | | | | | | | | | | | | | | |
| | Std. err. | 0.447 | 7.602 | 6.423 | 8.149 | 5.272 | 10.613 | | | | | | | | | | | | | | | | | | |
| | P-value | 0.071 | 0.085 | 0.362 | 0.142 | 0.630 | 0.958 | | | | | | | | | | | | | | | | | | |
| Promotion | Coef. | | | | | | | 3.288 | 11.784 | 6.784 | 30.501 | 4.705 | 21.126 | | | | | | | | | | | | |
| | Std. err. | | | | | | | 0.475 | 8.327 | 6.877 | 5.479 | 5.647 | 9.478 | | | | | | | | | | | | |
| | P-value | | | | | | | 0.000 | 0.175 | 0.336 | 0.000 | 0.414 | 0.041 | | | | | | | | | | | | |
| Positive impact | Coef. | | | | | | | | | | | | | -0.128 | 1.783 | 6.250 | 2.535 | 6.851 | -10.941 | | | | | | |
| | Std. err. | | | | | | | | | | | | | 0.498 | 9.265 | 7.155 | 9.710 | 5.741 | 11.072 | | | | | | |
| | P-value | | | | | | | | | | | | | 0.797 | 0.850 | 0.393 | 0.797 | 0.246 | 0.338 | | | | | | |
| Treatment 2 (Promotion, Positive impact, Avdanced forecast) | Coef. | | | | | | | | | | | | | | | | | | | 4.781 | -36.590 | 2.889 | -11.346 | 2.329 | -47.329 |
| | Std. err. | | | | | | | | | | | | | | | | | | | 0.779 | 11.842 | 12.027 | 10.371 | 9.599 | 16.516 |
| | P-value | | | | | | | | | | | | | | | | | | | 0.000 | 0.009 | 0.813 | 0.292 | 0.811 | 0.014 |
| Treatment 3 (Promotion, Negative impact, Baseline forecast) | Coef. | | | | | | | | | | | | | | | | | | | 4.496 | 1.350 | -9.452 | -8.360 | -9.104 | -1.174 |
| | Std. err. | | | | | | | | | | | | | | | | | | | 0.740 | 10.468 | 12.541 | 10.043 | 9.354 | 14.581 |
| | P-value | | | | | | | | | | | | | | | | | | | 0.000 | 0.899 | 0.462 | 0.419 | 0.344 | 0.937 |
| Treatment 4 (Promotion, Negative impact, Advanced forecast) | Coef. | | | | | | | | | | | | | | | | | | | -1.118 | -30.658 | -2.712 | -8.343 | -8.365 | -9.170 |
| | Std. err. | | | | | | | | | | | | | | | | | | | 0.713 | 9.535 | 11.352 | 9.233 | 9.061 | 13.751 |
| | P-value | | | | | | | | | | | | | | | | | | | 0.117 | 0.007 | 0.814 | 0.381 | 0.369 | 0.517 |
| Treatment 5 (Non-promotion, Positive impact, Baseline forecast) | Coef. | | | | | | | | | | | | | | | | | | | -2.471 | -33.532 | -7.020 | -43.676 | -8.791 | -46.258 |
| | Std. err. | | | | | | | | | | | | | | | | | | | 0.809 | 11.531 | 12.833 | 10.211 | 10.243 | 16.078 |
| | P-value | | | | | | | | | | | | | | | | | | | 0.002 | 0.012 | 0.592 | 0.001 | 0.403 | 0.014 |
| Treatment 6 (Non-promotion, Negative impact, Baseline forecast) | Coef. | | | | | | | | | | | | | | | | | | | -0.372 | -33.266 | -9.102 | -33.846 | -7.380 | -35.762 |
| | Std. err. | | | | | | | | | | | | | | | | | | | 0.726 | 10.070 | 11.530 | 9.329 | 9.207 | 14.029 |
| | P-value | | | | | | | | | | | | | | | | | | | 0.609 | 0.006 | 0.441 | 0.003 | 0.434 | 0.026 |
| **Random effects** | | | | | | | | | | | | | | | | | | | | | | | | | |
| Participant | Estimate | 1.372 | | | | | | 1.557 | | | | | | 1.382 | | | | | | 1.820 | | | | | |
| | Std. err. | 0.933 | | | | | | 0.945 | | | | | | 0.935 | | | | | | 0.950 | | | | | |
| Residual | Estimate | 133.346 | | | | | | 131.140 | | | | | | 133.488 | | | | | | 126.062 | | | | | |
| | Std. err. | 3.604 | | | | | | 3.544 | | | | | | 3.608 | | | | | | 3.410 | | | | | |
| Likelihood Ratio test | | 2.8 | | | | | | 3.65 | | | | | | 2.83 | | | | | | 5.21 | | | | | |
| p-value | | 0.047 | | | | | | 0.028 | | | | | | 0.046 | | | | | | 0.011 | | | | | |

*Note:* This table presents separate mixed-effects/regression models for each forecaster type under each hypothesis, with the exception of Hypothesis 1 (H1). For H1, which directly compares forecast accuracy between LLMs and humans, no separate analysis (e.g. H1a) is conducted for each forecaster type within this table.

**APPENDIX F**: Mean, Median, and Standard Deviation of Responses to the Post-Experimental Questionnaire

| Question | Mean | Median | Std. Dev. |
|---|---|---|---|
| **Q1-The importance of the following information in supporting your judgement when presented with <u>Baseline</u> statistical forecast (1 - Not important, 7 - Very important):** | | | |
| Forthcoming promotion type | 5.50 | 6 | 1.52 |
| Forthcoming external impact | 5.18 | 6 | 1.63 |
| The uplift size of previous promotions in general | 5.33 | 5 | 1.36 |
| The uplift size of previous similar promotions | 5.50 | 6 | 1.38 |
| Time series (sales) history | 5.50 | 6 | 1.41 |
| Baseline statistical forecast | 5.30 | 6 | 1.41 |
| Frequency of promotions | 5.11 | 5 | 1.44 |
| | | | |
| **Q2- The importance of the following information in supporting your judgement when presented with <u>Advanced</u> statistical forecast (1 – Not important, 7 – Very important):** | | | |
| Forthcoming promotion type | 5.68 | 6 | 1.26 |
| Forthcoming External impact | 5.46 | 6 | 1.41 |
| The uplift size of previous promotions in general | 5.41 | 6 | 1.23 |
| The uplift size of previous similar promotions | 5.59 | 6 | 1.25 |
| Time series (sales) history | 5.58 | 6 | 1.43 |
| Advanced statistical forecast | 5.82 | 6 | 1.17 |
| Frequency of promotions | 5.20 | 5 | 1.39 |
| | | | |
| **Q3- To what extent do you agree with the following statements (1 - Strongly disagree, 7 - Strongly agree):** | | | |
| The earlier promotions had a greater impact on your judgment about the forecast in period 25 | 5.02 | 5 | 1.45 |
| The later promotions had a greater impact on your judgment about the forecast in period 25 | 5.23 | 5 | 1.23 |
| All the promotions had an approximately the same impact on your judgment about the forecast in period 25 | 4.91 | 5 | 1.40 |
| In my forecasts, I considered only those periods of the provided sales time series that reflected similar conditions to the forecast period (for example, upward or downward trend) | 4.98 | 5 | 1.35 |
| In my forecasts, I considered all periods of the provided sales time series regardless of their similarity to the conditions of the forecast period | 5.10 | 5 | 1.33 |
| In anticipation of the objective feedback on your performance, how would you assess the accuracy of your judgemental forecasts made during the experiment? (1 - Very low, 7 - Very high) | 4.44 | 4 | 1.13 |